\documentclass[journal]{IEEEtran}
\usepackage{cite}
\pdfoutput=1
\ifCLASSINFOpdf
\else
\fi
\usepackage{amsmath,graphicx,cite,bm,booktabs,multirow,url}
\usepackage{algorithmic}
\usepackage{array}
\usepackage[marginal]{footmisc}
\usepackage{subfigure}
\usepackage{pdfpages}
\usepackage{color}

\begin{document}
\title{Competing Ratio Loss for Discriminative Multi-class Image Classification}
\author{Ke~Zhang,
        Yurong~Guo,
        Xinsheng~Wang,
        Dongliang~Chang,
        Zhenbing~Zhao,
        Zhanyu~Ma,
        and Tony X. Han% <-this % stops a space

\thanks{K. Zhang,~X. Wang and Z. Zhao are with Department of Electronic and Communication Engineering, North China Electric Power University, Hebei, China.}% <-this % stops a space
\thanks{K. Zhang,~ Y. Guo, ~D. Chang and Z. Ma are with Beijing University of Posts and Telecommunications, Beijing, China.}% <-this % stops a space
\thanks{T. X. Han is with Jingchi.ai, Guangzhou, China.}
\thanks{Corresponding author: Zhanyu Ma}
\thanks{This work is supported in part by the National Key R$\&$D Program of China under Grant 2019YFF0303300 and Subject II No. 2019YFF0303302; by National Natural Science Foundation of China (62076093, 61871182, 61922015, 61773071, and U19B2036); by Natural Science Foundation of Hebei Province (F2020502009); by Natural Science Foundation of Beijing (4192055, Z200002); by the Foundation Research Funds for the Central Universities (2020YJ006, 2020MS099); by the Beijing Academy of Artificial Intelligence (BAAI) (BAAI2020ZJ0204); by the Beijing Nova Programme Interdisciplinary Cooperation Project (Z191100001119140).}
}
\maketitle
\begin{abstract}
The development of deep convolutional neural network architecture is critical to the improvement of image classification task performance. Many image classification studies use deep convolutional neural network and focus on modifying the network structure to improve image classification performance. Conversely, our study focuses on loss function design. Cross-entropy Loss (CEL) has been widely used for training deep convolutional neural network for the task of multi-class classification. Although CEL has been successfully implemented in several image classification tasks, it only focuses on the posterior probability of the correct class. For this reason, a negative log likelihood ratio loss (NLLR) was proposed to better differentiate between the correct class and the competing incorrect ones. However, during the training of the deep convolutional neural network, the value of NLLR is not always positive or negative, which severely affects the convergence of NLLR. Our proposed competing ratio loss (CRL) calculates the posterior probability ratio between the correct class and the competing incorrect classes to further enlarge the probability difference between the correct and incorrect classes. We added hyperparameters to CRL, thereby ensuring its value to be positive and that the update size of backpropagation is suitable for the CRL’s fast convergence. To demonstrate the performance of CRL, we conducted experiments on general image classification tasks (CIFAR10/100, SVHN, ImageNet), the fine-grained image classification tasks (CUB200-2011, and Stanford Car), and the challenging face age estimation task (using Adience). Experimental results showed the effectiveness and robustness of the proposed loss function on different deep convolutional neural network architectures and different image classification tasks.  Code is released at https://github.com/guoyurong0104/CRL-code.
\end{abstract}

\begin{IEEEkeywords}
Competing ratio loss, cross-entropy loss, deep convolutional image classification, neural networks
\end{IEEEkeywords}
\IEEEpeerreviewmaketitle
\section{Introduction}
\label{sec:intro}
\IEEEPARstart{T}{he}  deep convolutional neural network (DCNN) has achieved great success in the most tasks of computer vision, such as image classification\cite{resnet,huang2017densely}, object detection\cite{fasterrcnn,focalloss}, image segmentation\cite{segnet}, a DCNN can automatically learn the optimal feature from inputs in an end-to-end manner. As the basis of computer vision technologies, image classification has been a research focus in artificial intelligence fields  \cite {Lecun2015Deep, gu2018recent, 8447427, 8454883, 8730301, 8666165}. 
\subsection{Effects of Other Methods, Programs and Loss Function on DCNN Classification}

In the image classification task, DCNNs learn to generate the predicted distribution of image class by extracting the features of an input image\cite{deeplearning}. The learning objective is to minimize the difference between the class distribution predicted by DCNNs and true data-generating distribution. To measure the difference, many loss functions have been proposed, such as mean square error loss (MSE) \cite{mse}, hinge loss\cite{hingeloss} and cross-entropy loss (CEL)\cite{celoss}. These loss functions play important roles during the training of DCNNs. Compared with MSE and hinge loss, CEL has excellent convergence speeds for training DCNNs \cite{convergence}. Therefore, CEL is a reasonable loss function for classification tasks based on DCNNs.

Even though CEL is commonly used for image classification tasks, it also has clear disadvantages. The cross entropy between two probability distributions over the same set of events measures the average number of bits needed to identify an event drawn from the set if the coding scheme obeyed a learned probability distribution, rather than the real but unknown distribution\cite{celoss}; however, in practice, i.e., when training DCNNs in the classification task, the real data-generating distribution is unknown and replaced by the empirical probability distribution over a training set. Each sample of the training set is independently and identically distributed (i.i.d) from the data space \cite{vapnik2013nature}. Under the assumption that both the image feature space and the label space obey uniform distribution, minimizing CEL is equivalent to maximizing the likelihood. In other words, minimizing the CEL in training DCNNs is equivalent to maximizing the likelihood of the training samples \cite{deeplearning}. Maximum likelihood is a training criterion of machine learning, by which the network learns the likelihood of a correct class for an input sample. The network uses Bayes rules to calculate posterior probabilities of target classes for the input sample and then predicts the most likely class. We observed that when the labels of training samples are one-hot, this criterion cannot directly discriminate the posterior probability of a correct class against the classes belonging to the competing incorrect classes, as it only focuses on the posterior probability of a correct class. Zhu et al.\cite{zhu2018negative} introduced the probability of incorrect classes into Negative Log Likelihood Ratio Loss (NLLR) to better discriminate the correct class from the competing incorrect classes. Generally speaking, a loss function plays an important role as it must faithfully distill all the aspects of the model down into a single number in such a way that improvements in that number are a sign of a better DCNN. Optimization of the loss function is normally presented as a minimization problem\cite{deeplearning}. However, when using NLLR for training DCNNs, the value of NLLR is not constantly positive or negative. The value of the NLLR for training DCNN significantly affects the training results.

\begin{figure}[]
	\begin{minipage}[b]{1.0\linewidth}
		\centering
		\centerline{\includegraphics[width=8.5cm]{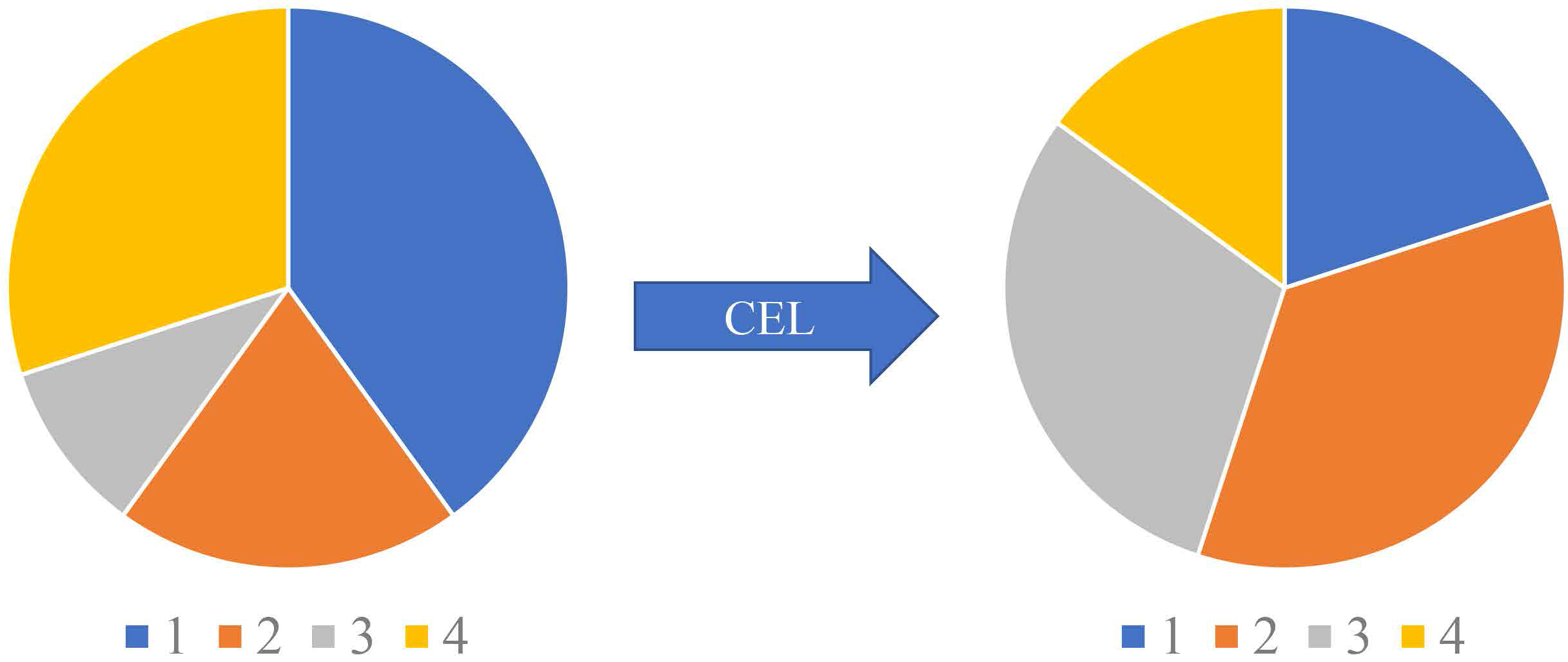}}
		%  \vspace{2.0cm}
		\centerline{(a) Cross Entropy Loss}\medskip
		\label{motivationa}
	\end{minipage}
	\hfill
	\begin{minipage}[b]{1.0\linewidth}
		\centering
		\centerline{\includegraphics[width=8.5cm]{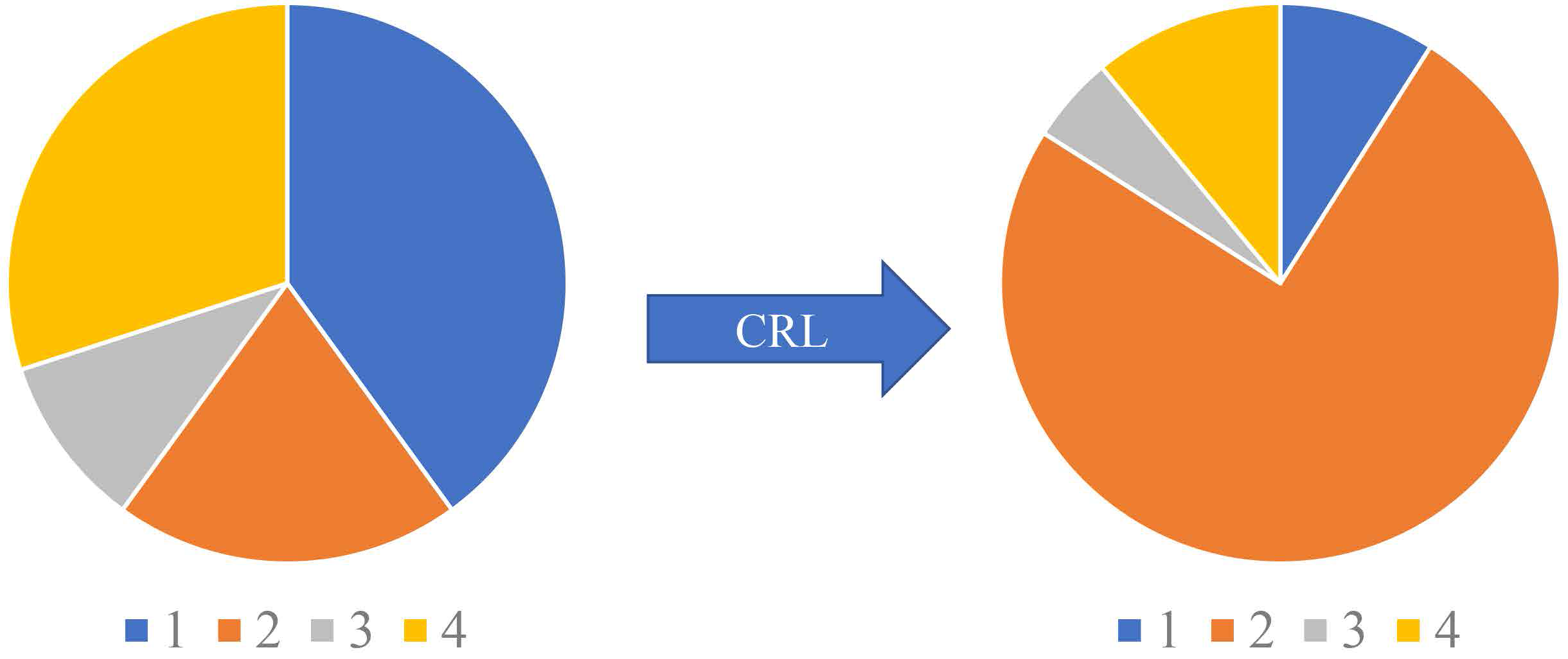}}
		%  \vspace{1.5cm}
		\centerline{(b) Competing Ratio Loss}\medskip
		\label{motivationb}
	\end{minipage}
	\caption{Motivation of Competing Ratio Loss. The area of each sector represents the probability of each class. 
		The left and right pie all represents the softmax outputs of image using loss. And when training classifier, it changed from the left pie to right pie.
		Class 2 is the correct class. (a) An illustration of cross entropy loss function, which focuses on the posterior probability of correct class when the labels of training images are one-hot. It cannot be discriminative against the incorrect classes. When training classifier, the posterior probability of some incorrect class (e.g. Class 3) may be probably increased. (b) The idea of the competing ratio loss, which introduces competing ratio between the probability of the correct class (Class 2) and the probability of the competing incorrect classes (Class 1, 3 and 4), increases the posterior probability of the correct class, and decreases the posterior probabilities of the incorrect classes. It enlarges the probability difference between the correct and incorrect classes. The correct class can be easily distinguished from the competing incorrect ones.}
	\label{motivation}
\end{figure}

The NLLR/DCCN problem led us to propose the competing ratio loss (CRL), which aims to discriminate the posterior probability of a correct class against the competing incorrect classes. Moreover, the value of CRL guarantees that the optimization of the loss function is a minimization problem, which better enlarges the probability difference between the correct and incorrect classes, as shown in  \textbf{Fig.\ref{motivation}}. 

\subsection{Contributions of This Research to the Computer Science Research Community}

\subsubsection{A new loss function for image classification}The competing ratio loss (CRL), was developed to discriminate the posterior probability of correct class against the competing incorrect classes. When training DCNNs, the difference between the negative log likelihood of the correct class and the negative log likelihood of competing incorrect classes increases, and the value of CRL is positive during the training procedure of DCNN, ensuring the parameters of DCNN is updated towards the negative gradients in each iteration, which means CRL can converge to the local optimum faster. Moreover, we add a regularization term to the cross-entropy loss, which can alleviate the vanishing of the gradient when the value of the cross-entropy loss is close to zero. 

 \subsubsection{More flexibility and good versatility}The proposed CRL can be embedded into different types of DCNNs, such as ResNet\cite{resnet}, VGGNet\cite{vgg},  DenseNet\cite{huang2017densely}, and MobileNetV2\cite{Sandler_2018_CVPR} and optimized via the stochastic gradient descent (SGD) in an end-to-end manner. In addition, our proposed CRL was evaluated on a number of image classification datasets of different aspects (the general image classification datasets, including CIFAR-10/100\cite{cifar}, SVHN\cite{SVHN} and ImageNet\cite{imageNet}, as well as the fine-grained image classification datasets including CUB200-2011\cite{CUB200-2011} and Stanford Car\cite{STCAR}, and the challenging face age estimation dataset including Adience\cite{Adience}). CRL achieved better results than CEL and NLLR, which implies that CRL has good flexibility and versatility in different types of image classification tasks. 

We note that a shorter conference version of this manuscript appeared in Zhang and Wang\cite{crl}. Our initial conference paper did not address the problem of CRL's effectiveness on complex image classification datasets. This manuscript addresses this issue and provides additional analysis on the convergence speed of the proposed CRL.

\section{Related work}
\label{sec:Related works}

The performance improvement of image classification cannot be separated from the development of the structure and depth of deep convolutional neural networks.  Since AlexNet\cite{alexnet} won the Large Scale Visual Recognition Challenge of 2012 (ILSVRC2012), which represented a significant advancement in image classification, a lot of deep neural networks have been proposed. The VGGNet\cite{vgg} improved performance by deepening the depth of the network. GoogLeNet\cite{googlenet,inception-v2,inception-v3,inception-v4,xception} enhanced the image feature extraction capabilities by using multiple scale convolution kernels on a single-layer convolutional layer. To reduce the difficulty of training deep convolutional neural networks, He et al. proposed residual network (ResNet)\cite{resnet}, which introduced shortcut connections and residual representation during training network training. Thus, ResNet has several popular variants, which explore the representation ability of DCNNs. Wide residual networks (WRNs) \cite{WRN} widened the network by increasing the number of output channels in the convolutional layer. ResNeXt\cite{ResNext} increased the third dimensional–cardinality to improve image classification performance. Zhang et al.\cite{ror} proposed multilevel residual networks of residual networks, i.e., the Residual Networks of Residual Networks (RoR), which added level-wise shortcut connections upon original residual networks to promote the learning capability of residual networks. Then, they built a pyramidal multilevel residual network (P-RoR) \cite{pror} based on the pyramid residual network. To ensure a maximum information flow between layers in the networks, Huang et al. \cite{huang2017densely} proposed DenseNet, wherein each layer obtains additional inputs from all preceding layers and passes on its own feature maps to all the subsequent layers. CondenseNet\cite{condensenet} was proposed to reduce the memory of DenseNet by learning group convolution operations and pruning during training. In addition, the dual path network (DPN) \cite{DPN} family combined ResNet, which enabled feature reusage, and DenseNet, which enabled new feature exploration. These innovations achieved competitive results in image classification, object detection, and semantic segmentation tasks. Alternately, updated clique (CliqueNet) models\cite{cliquenet}  incorporate both forward and backward connections between any two layers in the same block. These connections maximize information flow and achieve feature refinement. Zhang et al. proposed a multiple feature reweight DenseNet (MFR-DenseNet) \cite{mfrdensenet} to improve the representation power of DenseNet by adaptively recalibrating the channel-wise feature responses and explicitly modeling the interdependencies between the features of different convolutional layers.

However, after the network’s depth, width and parameter quantities reach a certain maximum level; hence, even if the network’s depth is further increased, the performance of the classifier will not be greatly improved, but a large amount of computing resources will be consumed. Thus, some researchers turn their attention to other aspects of convolutional neural networks such as loss function optimizers. To train the classification network, cross entropy loss (CEL) is by far the most popular loss function. In addition, some existing works have attempted to improve the original CEL from different aspects. Triplet Loss\cite{tripletloss} focuses on reducing the distance between the current sample and a positive sample, and increases the distance for the negative ones. Center loss\cite{centerloss} and increases the distance for the negative ones. Center loss [37] was proposed to simultaneously learn a center for the deep features of each class and constrain the distances between the deep features and their corresponding class centers, which reduced intra-class variance. Liu et al. \cite{largemarginloss} proposed large-margin loss (L-Softmax) by adding angular constraints to each identity to encourage the discriminative learning of features by increasing inter-class separability and intra-class compactness. Large-margin losses have a lot of variant functions, such as Soft-Margin Softmax Loss\cite{softmarginsoftmax}, Angular Softmax (A-Softmax)\cite{A-Softmaxloss} and so on. Soft-Margin Softmax Loss (SM-Softmax) uses a soft distant margin to theoretically contain all the hard angle margin in L-Softmax and the degenerative margin in Softmax Loss. SM Softmax not only inherited all the Softmax and L-Softmax merits but also learned features with a large soft margin between different classes. A-Softmax\cite{A-Softmaxloss} improved L-Softmax by normalizing the weights, which achieved a better performance on a series of open-set face recognition benchmarks. Li et al.\cite{focalloss} proposed focal loss (FL) to address this class imbalance by reshaping the standard CEL such that it down-weighted the loss assigned to well-classified examples. Li et al.\cite{DualCE} proposed dual cross-entropy loss (DualCE) to address this class imbalance by reshaping the standard CEL such that it down-weighted the loss assigned to well-classified examples. Li et al.\cite{tamingloss} proposed taming cross entropy loss (TCE), which is more robust to noise and outliers. 

\section{Competing Ratio Loss Function}
\label{sec:Competing Ratio Loss Function}
\subsection{Cross Entropy Loss Function}
\label{ssec:cross entropy loss function}
The goal of the image classification task is to use DCNN to predict the class of a given image. During this progress, cross-entropy loss function (CEL) is the most common used loss function. CEL can measure the performance of a classification model, representing the difference between target class distribution and predicted class distribution. Besides, common optimization algorithms such as stochastic gradient descent (SGD) can effectively reduce the value of CEL.

Assumed all $N$ images have $C$ classes in the image classification task. The input of DCNN is an image $x_i$ of all $N$ images, DCNN predicts $C$ nodes, each of which represents the score of the class corresponding to image $x_i$. Under the premise that the activation function of DCNN is softmax function, the output of C nodes mimics the posterior probabilities of the class corresponding to image $x_i$: $\hat{p}(y_c|\bm{x})$, $c=1,2,\cdots,C$, 
where \bm{$x$} is the input vector of DCNN, $x_i \in \bm{x}$,$ i=1,2,\cdots,N$. 

Therefore, CEL is defined as:
\begin{equation}
\label{ce}
L_{CEL}=-\sum\limits_{i=1}^{N}{\sum\limits_{c=1}^{C}{p(y_{c}^{{}}|\bm{x})}\log \hat{p}\left( y_{c}^{{}}|\bm{x} \right)},
\end{equation}
where $p(y_c|\bm{x})$ is the empirical distribution of the training set, and $\hat{p}(y_c|\bm{x})$ is the predicted distribution from the DCNN. In the specific (and usual) case of multi-class image classification, each training image is labeled based on the correct class it belongs to:

\begin{equation}
\label{condition}
p(y_c|\bm{x})=
\begin{cases}
1& \text{if }  \bm{x}\in y_c\\
0& \text{otherwise}.
\end{cases}
\end{equation}

Based on Equation (\ref{condition}), Equation (\ref{ce})can be rewritten:

\begin{equation}
\label{rece}
L_{CEL}=-\sum\limits_{i=1}^{N}{\log{\hat{p}(y_c|\bm{x})}}.
\end{equation}

Equation (\ref{rece}) shows that when the training label images are one-hot, CEL only focuses on the probability that an image is assigned to its ground-truth class and does not place any focus on the probability that the image is assigned to a class other than its ground-truth class (competing incorrect class), as shown in  \textbf{Fig.\ref{motivation}(a)}. \textbf{Fig.\ref{motivation}(a)} shows that the probability distribution change of the prediction when minimizing CEL. The figure shows the probabilities of some classes increasing and the probabilities of some classes decreasing, except for the probability of the correct class. CEL cannot directly discriminate the posterior probability of the correct class against the competing incorrect classes. 

When $C=2$, the image classification task turns into the binary image classification task, which answers the question of whether the image $x_i$ belongs to a certain class. During the progress, the binary classifier based on DCNN can only have one single output for which the empirical probability $p(y_c|\bm{x})$ equals 1 for one class and 0 for the other class. Then, CEL can be simplified to a binary cross-entropy loss function (BCE) as:

\begin{equation}
\label{bce}
\begin{split}
L_{BCE}=-\sum\limits_{i=1}^{N}{\{p(y_c|\bm{x})\log{\hat{p}(y_c|\bm{x})}}\\
+[1-p(y_c|\bm{x})]\log{[1-\hat{p}(y_c|\bm{x})]\}}.
\end{split}
\end{equation}

When $C\ge3$, the image classification task turns into the multi-class image classification task. According to Equation (\ref{condition}), the binary cross-entropy loss function can be extended to categorical cross entropy (CCE) loss for one-hot encoding as: 
\begin{equation}
\label{rebce}
\begin{split}
{L}_{CCE}=-\sum\limits_{i=1}^{N}{\{\log \hat{p}({{y}_{c}}|\bm{x})+\sum\limits_{k=1,k\ne c}^{C}{\log [1-\hat{p}({{y}_{k}}|\bm{x})]\}}}\\ 
=-\sum\limits_{i=1}^{N}{\{\log [\hat{p}({{y}_{c}}|\bm{x})\prod\nolimits_{k=1,k\ne c}^{C}{\hat{p}(\bar{y_k}|\bm{x})}]\}},
\end{split}
\end{equation}
where $\hat{p}(\bar{y_k}|\bm{x})=1-\hat{p}(y_k|\bm{x})$ represents the probability of $\bm{x}$ not predicted as $y_k$, in other words, the predicted probability of $x$ belongs to the competing incorrect classes.  Equation (\ref{rebce}) shows that in CCE, each $\hat{p}({y_c}|\bm{x})$ is multiplied by a factor, which is the product of probabilities of belonging to each of the competing incorrect classes. Although CCE consists of probabilities pertaining to competing incorrect classes, CCE cannot also discriminate the correct class probability from the competing incorrect ones.

Notably, CEL can be backpropagated when training DCNNs. Assume that a softmax layer is connected with DCNN. We denote $\hat{p}(y_c|\bm{x})$ as $p_c$, $\hat{p}(y_k|\bm{x})$ as $p_k$. $p_c$ and $p_k$ are outputs of the softmax layer. The softmax function guarantees ${{p}_{c}}+\sum\nolimits_{k=1,k\ne c}^{C}{p_{k}^{{}}}=1, {{p}_{c}}\in (0,1)$. Therefore (\ref{rece}) can be written as
\begin{equation}
\label{recece}
L_{CEL}=-\log{p_{c}}.
\end{equation}

The softmax layer outputs $p_c$. The derivative of $p_c$ with respect to $x_j$ in the  softmax function is:
\begin{equation}
\frac{\partial p_{c}^{{}}}{\partial x_{j}^{{}}}=
\begin{cases}
p_{c}^{{}}(1-p_{j}^{{}})&   c\ne j\\
-p_{c}^{{}}p_{j}^{{}}& c=j.
\end{cases}
\end{equation}

Therefore the derivative of CEL with respect to $x_j$ is:
\begin{equation}
\label{daoshuce}
\begin{split}
\frac{\partial L_{CEL}^{{}}}{\partial x_{j}^{{}}}=\frac{\partial L_{CEL}^{{}}}{\partial p_{c}^{{}}}\times \frac{\partial p_{c}^{{}}}{\partial x_{j}^{{}}}=\\
(-\frac{1}{p_{c}^{{}}})\times (-p_{c}^{{}}p_{j}^{{}})=p_{j}.
\end{split}
\end{equation}

Thus, (\ref{daoshuce}), confirms that the gradient that CEL updates when backpropagating is not relevant to $p_{c}$.

\subsection{Negative Log Likelihood Ratio Loss and its weakness}
\label{ssec:Negative Log Likelihood Ratio Loss}

Section \textbf {\ref{ssec:cross entropy loss function}} analyzes CEL and its convergence. CEL only measures the probability of the correct class for each input image, which cannot learn to discriminate the probability between the correct class and competing incorrect class probabilities. Zhu et al. \cite{zhu2018negative} proposed a negative log likelihood ratio loss (NLLR) to directly discriminate the correct class probability from the competing incorrect ones, defined as:

\begin{align}
\label{nllr}
L_{NLLR}&=-\log{\frac{\hat{p}(y_c|\bm{x})}{\sum\limits_{k=1,k\ne c}^{C}{\hat{p}(y_k|\bm{x})}}.}
\end{align}

Equation (\ref{nllr}) calculates the ratio between the predicted correct-class probability and the probabilities of competing incorrect classes to widen the probability difference between the correct and competing incorrect classes.

$\hat{p}(y_k|x)$ is denoted as $p_k$, $\hat{p}(y_c|x)$ as $p_c$. $p_k$ and $p_c$ are the outputs of softmax layer. The softmax function guarantees ${{p}_{c}}+\sum\nolimits_{k=1,k\ne c}^{C}{p_{k}^{{}}}=1,{{p}_{c}}\in (0,1)$. Therefore (\ref{nllr}) can now be written as:
\begin{align}
\label{cr}
L_{NLLR}&=\log{\frac{1-p_c}{p_c}}\notag\\
&=\log{(1-p_c)}-\log{p_c}\\
&=\log{(1-p_c)}+L_{CEL}.\notag
\end{align}

Obviously, when $p_c\in(0,0.5)$, we have $\log{\frac{(1-p_c)}{p_c}}>0$. When $p_c\in(0.5,1)$, we obtain $\log{\frac{(1-p_c)}{p_c}}<0$. This demonstrates that the value of NLLR is not always positive or negative when training DCNN. Generally speaking, optimization of the loss function is normally presented as a minimization problem\cite{deeplearning}. The value of NLLR in training DCNN may severely affect the convergence of NLLR.

\subsection{Competing Ratio Loss Function}
\label{ssec:competive ratio loss function}

To solve the problem minimization problem in connection with  Equation (\ref{cr}), we propose the competing ratio loss (CRL) as:

\begin{align}
\label{recr}
L_{CRL}&=\log{\frac{(\alpha+\sum\limits_{k=1,k\ne c}^{C}{\hat{p}(y_k|\bm{x})})^\beta}{\hat{p}(y_c|\bm{x})}}\notag\\
&=\beta\log{(\alpha+\sum\limits_{k=1,k\ne c}^{C}{\hat{p}(y_k|\bm{x})})}-\log{\hat{p}(y_c|\bm{x})}\notag\\
&=\beta\log{(\alpha+1-p_c)}+L_{CEL}.
\end{align}

Equation (\ref{recr}) manifests that CRL is a combination of CEL and a regularization term, which is responsible for modifying the probability ratio between the predicted correct-class and competing incorrect classes.  The hyper-parameter $\alpha\geq1$ guarantees the loss value to be constantly positive; namely, the optimization of CRL is a minimization problem wherein CRL has a better convergence. The weight of the competing ratio is modified by the hyperparameter $\beta$. When $\beta=0$, CRL is equivalent to CEL. It is illustrated that the probability ratio between the correct and competing incorrect-class probabilities decreases with increase of the posterior probability of the correct class, which results in widening the probability difference between the correct class and competing incorrect classes 

Additionally, the numerator of Equation (\ref{recr}) can be simplified the sum of competing class probabilities representing the probability of not belonging to the correct class, denoted as $\hat{p}(\bar{y_c}|\bm{x})$. Supposed the image feature distribution $p(x)$ and the class label distribution $p(y)$ are not relevant to the CNN parameters and obeying uniform distributions. Thereby, according the Bayesian inference, Equation (\ref{rece}) and Equation (\ref{recr}) can be rewritten as 
\begin{equation}
\label{recebi}
L_{CEL}=-\log{\hat{p}(\bm{x}|y_c)}.
\end{equation}
\begin{align}
\label{recerbi}
L_{CRL}&=\beta\log{(\alpha+\hat{p}(\bar{y_c}|\bm{x}))}-\log{\hat{p}(y_c|\bm{x})}\notag\\
&=\beta\log{(\alpha+\hat{p}(\bm{x}|\bar{y_c}))}-\log{\hat{p}(\bm{x}|y_c)}\\
&=-[\log{\hat{p}(\bm{x}|y_c)}-\beta\log{(\alpha+\hat{p}(\bm{x}|\bar{y_c}))}].\notag
\end{align}

Equation (\ref{recebi}) shows that CEL is the negative log likelihood of image $\bm{x}$. Equation (\ref{recerbi}) shows that CRL can be considered as an approximation of the negative log likelihood difference between the correct class and competing incorrect classes. It directly discriminates the correct class from the competing incorrect classes for each training image when training DCNN. 

\subsection{Convergence of the Competing Ratio Loss Function}
\label{ssec:derivative of the competing ratio loss function}
$\hat{p}(y_c|\bm{x})$ is denoted as $y_c$. The softmax layer outputs $p_c$. The derivative of CRL with respect to $x_j$ is
\begin{equation}
\label{daoshu}
\begin{split}
\frac{\partial L_{CRL}^{{}}}{\partial x_{j}^{{}}}&=\frac{\partial L_{CRL}^{{}}}{\partial p_{c}^{{}}}\times \frac{\partial p_{c}^{{}}}{\partial x_{j}^{{}}}\\
&=(\frac{\beta}{\alpha +1-p_{c}^{{}}}-\frac{1}{p_{c}^{{}}})\times (-p_{c}^{{}}p_{j}^{{}})\\
&=(1-\frac{\beta p_{c}}{\alpha+1-p_{c}})p_{j}.
\end{split}
\end{equation}

\begin{figure}[!t]
	\centering
	\includegraphics[width=0.5\textwidth]{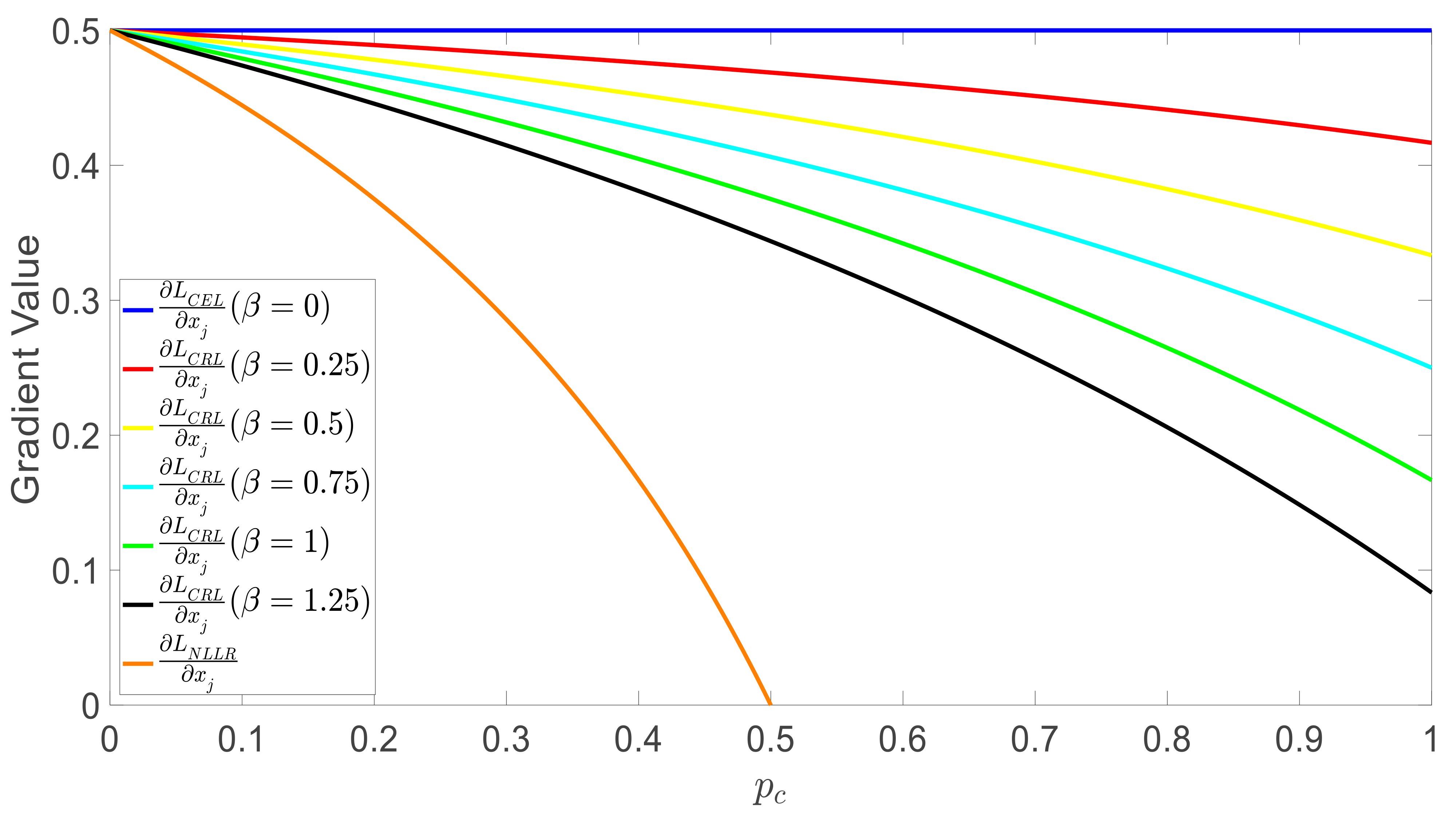}
	\caption{The trend of CEL, NLLR and CRL's gradient value against $x_j$ with $p_c$.  The blue line shows the trend of CEL gradient value against $x_j$ with $p_c$. The orange line shows the trend of NLLR gradient value against $x_j$ with $p_c$. The others line show the trend of CRL gradient value against $x_j$ with $p_c$ when $\beta$ takes different values.}
	\label{fig:gradient}
\end{figure}

Equation (\ref{daoshu}) shows the derivative of CRL can be propagated using the standard back propagation method. Compared (\ref{daoshu}) and (\ref{daoshuce}), we can observe that the differentiation of CRL is relevant to $p_{c}$.  \textbf{Fig. \ref{fig:gradient}} shows the trend of CRL and CEL's gradient value against $x_j$ with $p_c$. For the convenience of drawing the figure, $\alpha=1.5$ and $p_j=0.5$ in (\ref{daoshuce}) and (\ref{daoshu}). When $p_c$ increases, the gradient value of CRL relative to $x_j$ is decreasing. However, the gradient value of CEL relative to $x_j$ is constant with respect to $x_j$. Thus, CRL's update step size in SGD is dependent of the confidence estimate of the correct class in the classification prediction. When $p_c$ increases, the update step size is decreasing. This illustrates that when a CNN classifier has a higher probability of the correct class, the parameter amount of CNN updated by backpropagation becomes smaller, maintaining high confidence in the correct class of images by the CNN classifier. On the contrary, when $p_c$ decreases, the update step size is increasing. The CNN classifier has higher probabilities of the incorrect classes, the parameter amount of CNN updated by backpropagation becomes larger, making CNN tend to increase the probability of the correct class. This indicates that the CRL is adaptive to the probability of correct class selection in backpropagation. When the probability of correct class is high, CNN updates a small number of parameters. When the probability of correct class is low, CNN updates a large number of parameters to increase the probability of correct class. Compared with CRL, the update size of backpropagation using CEL is constant whether the probability of the correct class is high or low. CEL is unchanged for the probability of the correct class. Moreover, the hyperparameter $\beta$ can influence the update step size of the CRL, as shown in \textbf{Fig.\ref{fig:gradient}}. From (\ref{daoshu}), we can obviously conclude that when $\beta$ increases, the gradient value of CRL relative to $x_j$ decreases, maintaining a higher confidence in the correct classification of images during backpropagation.

Besides, according to (\ref{daoshu}), when $\alpha=0$, $\beta=1$, and NLLR is equivalent to CRL. The derivative of NLLR with respect to $x_j$ is:

\begin{equation}
\label{nllrdaoshu}
\begin{split}
\frac{\partial L_{NLLR}^{{}}}{\partial x_{j}^{{}}}&=\frac{\partial L_{NLLR}^{{}}}{\partial p_{c}^{{}}}\times \frac{\partial p_{c}^{{}}}{\partial x_{j}^{{}}}\\
&=(\frac{1}{1-p_{c}^{{}}}-\frac{1}{p_{c}^{{}}})\times (-p_{c}^{{}}p_{j}^{{}})\\
&=(1-\frac{p_{c}}{1-p_{c}})p_{j}.
\end{split}
\end{equation}

\textbf{Fig. \ref{fig:gradient}} also shows the trend of NLLR's gradient value against $x_j$ with $p_c$. When $p_c>0.5$, the gradient value of NLLR relative to $x_j$ is negative, which leads to  training DCNN by using, stochastic gradient descent (SGD); thus, the parameters of DCNN may be updated towards the positive gradients in each iteration, and NLLR cannot converge to local optimum fast. In contrast to NLLR, the gradient value of CRL relative to $x_j$ is constantly positive, assuring that the parameters of DCNN are updated towards the negative gradients in each iteration; thus, CRL can now converge to local optimum fast.

\section{Experiments and Results}
\label{sec:experiments}
In this section, we show implementation details and experimental results. First, we show how two hyperparameters  $\alpha$ and $\beta$   influence the CRL. Then we show the robustness and effectiveness of CRL in different-depth DCNNs and different types of DCNNs.
Finally, we show our CRL can achieve better generalization on different types of image classification tasks, including general image classification, challenging fine-grained image classification, and difficult-to-determine face age estimation.
\subsection{Experiment Setups}
\label{ssec:experiments setup}
\subsubsection{Datasets}

The experiments in this manuscript are performed on three different types of image classification tasks, including general image classification, challenging fine-grained image classification and hard face age estimation.

\paragraph{General image classification datasets}
\textbf{CIFAR10/100}\cite{cifar} consist of colored natural scene images, with 32$\times$32 pixels each. The training set and test set contain 50 000 and 10 000 images respectively. CIFAR-10 images are drawn from 10 classes, and the CIFAR-100 images are drawn from 100 classes. We adopt a standard data augmentation scheme in our experiments: random sampling and horizontal flipping. 

The Street View House Numbers (\textbf{SVHN})\cite{SVHN} dataset contains 32$\times$32 colored digit images. There are 73 257 images in the training set, 26 032 images in the test set, and 531 131 images for additional training. For training network, we use all the training data without any data augmentation.

The ILSVRC 2012 classification dataset (\textbf{ImageNet})\cite{imageNet} is the largest large-scale image classification dataset currently, consisting of 1.2 million images for training, and 50 000 for validation, from 1,000 classes. We adopt the same data augmentation scheme for training images as in \cite{huang2017densely}, and apply a single-crop with size 224 $\times$ 224 at test time.

\paragraph{Fine-grained image classification datasets}

CaltechUCSD Birds-200-2011 (\textbf{CUB-200-2011})\cite{CUB200-2011} is a challenging dataset, which aims to distinguish subordinate-level bird species, with photos of 200 bird species, each species with roughly 30 training images and 30 testing images. This dataset has become a staple for testing new ideas for fine-grained image classification. 

Stanford CAR (\textbf{STCAR}) \cite{STCAR} contains 16,185 images of 196 classes of cars. The data is split into 8,144 training images and 8,041 testing images, where each class has been split roughly in a 50-50 split.

\paragraph{Face age estimation datasets}

The \textbf{Adience}\cite{Adience} dataset is proposed for facilitating the study of age estimation in the wild, which is difficult to classify due to the similarity between adjacent age groups. The entire Adience collection includes 26 580 256 $\times$ 256 color facial images of 2 284 subjects, with eight classes of age groups (0-2, 4-6, 8-13, 15-20, 25-32, 38-43, 48-53, 60-100). In this article, testing for age group classification is performed using a standard five-fold, subject-exclusive cross-validation protocol, defined in\cite{Adience}.

\subsubsection{Network use and training strategies}

For comparison, the experiments in this manuscript use ResNet\cite{resnet}, VGG\cite{vgg},  DenseNet\cite{huang2017densely}, and MobileNetV2\cite{Sandler_2018_CVPR} networks. On CIFAR datasets, the networks are trained using stochastic gradient descent (SGD) for 300 epochs with a mini-batch size of 256. We use a weight decay of 1e-4, Nesterov momentum of 0.9. The learning rate starts from 0.1, and is divided by 10 at 50\% and 75\% of the training procedure. On SVHN, the networks are trained using SGD for 40 epochs with a mini-batch size of 64. The learning rate start from 0.1, and are divided by 10 at 50\% and 75\% of the training procedure. networks. On CIFAR datasets, the networks are trained using stochastic gradient descent (SGD) for 300 epochs with a minibatch size of 256. We use a weight decay of 1e-4 and a Nesterov momentum of 0.9. The learning rate starts at 0.1 and is divided by 10 at 50\% and 75\% of the training procedure. On SVHN, the network is trained using SGD for 40 epochs with a mini-batch size of 64. The learning rate start from 0.1 and is divided by 10 at 50\% and 75\% of the training procedure.

For the CUB200-2011 and Stanford CAR datasets, we used the same experiment setup as described in \cite{BCNN,DFLCNN,DCL} for verifying the generalization of CRL in a fine-grained image classification dataset. 

On the Adience dataset, to better learn the image features in Adience, we fine-tuned the ResNet model by pretraining on ImageNet. When we used the pretrained ResNet model to fine-tune on Adience, we replaced the 1 000 classes prediction layer with an 8-class age prediction layer. We used SGD with a mini-batch size of 64 for 120 epochs to fine-tune on Adience. The learning rate starts from 0.01 and is divided by a factor of 10 after epoch 80. 

On ImageNet, we trained models for 90 epochs with a batch size of 256. The learning rate is initially set to 0.1 and is lowered by 10 times at epoch 30 and 60. All experiments are implemented on Pytorch 1.0 with one NVIDIA GeForce GTX TITAN X Pascal GPU.

\subsection{Influences of the parameters $\alpha$ and $\beta$}
\label{ssec:Influences of the parameters alpha and beta}

There are two hyperparameters $\alpha$ and $\beta$. Since in (\ref{recr}), $\alpha$ guarantees that the loss function is greater than 0, and $\beta$ is the weight of the competing ratio, we evaluated their influences in image classification on CIFAR-10 and CIFAR-100 using the ResNet34\cite{resnet} model.
\begin{table}[]
	\centering
	\caption{Test Errors (\%) on CIFAR-10/100 under different hyperparameters $\alpha$ and $\beta$ by ResNet34}
	\begin{tabular}{|c|c|c|c|c|}
		\hline
		\textbf{CIFAR-10} & $\alpha$\textbf{=0} & $\alpha$\textbf{=1} & $\alpha$\textbf{=1.5} & $\alpha$\textbf{=2} \\ \hline \hline
		$\beta$\textbf{=0} & \multicolumn{4}{c|}{6.63} \\  \cline{2-5}
		$\beta$\textbf{=0.25}& 7.44 & 6.48 & 6.33 & 6.44 \\ 
		$\beta$\textbf{=0.5} & 7.79& 6.18 & 6.08 & 6.23 \\ 
		$\beta$\textbf{=0.75} & 8.31& 6.28 & 6.07 & 6.18 \\ 
		$\beta$\textbf{=1} & 8.55 & 6.41 & \textbf{5.99} & 6.1 \\ 
		$\beta$\textbf{=1.25} & 7.8 & 6.24 & 6.04 & 6.57 \\ \hline\hline
		\textbf{CIFAR-100} & $\alpha$\textbf{=0} & $\alpha$\textbf{=1} & $\alpha$\textbf{=1.5} & $\alpha$\textbf{=2} \\ \hline\hline
		$\beta$\textbf{=0} & \multicolumn{4}{c|}{27.87} \\ \cline{2-5}
		$\beta$\textbf{=0.25} & 28.04& 28.14 & 27.39 & 26.78 \\ 
		$\beta$\textbf{=0.5} & 28.79& 27.56 & 27.93 & 27.18 \\ 
		$\beta$\textbf{=0.75} & 30.01& 27.53 & 27.36 & 27.47 \\ 
		$\beta$\textbf{=1} & 29.67& 27.34 & \textbf{27.26} & 27.54 \\ 
		$\beta$\textbf{=1.25} & 29.78& 28.03 & 27.78 & 28.01 \\ \hline
	\end{tabular}
	\label{alphabeta}
\end{table}

\begin{table}[!t]
	\centering
	\caption{Test accuracies on CIFAR-10 under condition described in \cite{zhu2018negative}}
	\begin{tabular}{|c|c|c|c|}
		\hline
		\textbf{\#Epoch}& \textbf{NLLR} & \textbf{CEL} & \textbf{CRL}($\alpha=1.5,\beta=1$)\\ \hline\hline
		100  &80.19 & 83.85 & \textbf{84.29}\\ 
		500  &85.31 & 85.95 & \textbf{86.59}\\ \hline
	\end{tabular}
	\label{nllr2}
\end{table}

We set $\alpha$ four values: 0, 1.0, 1.5 and 2.0, and we set $\beta$ five values: 0, 0.25, 0.5, 0.75 and 1. When $\alpha=0$, $\beta=1$, and CRL is equivalent to NLLR. \textbf{Table \ref{alphabeta}} shows the influence of hyperparameters $\alpha$ and $\beta$. It can be observed that when $\alpha$= 0, the test errors are worse than the results in condition $\alpha>0$. In addition, 
we can see that when $\alpha=0$ and $\beta \ne 1$, the test errors are much worse than the results in condition  $\beta=0$ or $\alpha > 0$. And when $\alpha > 0$, the test errors only fluctuate slightly. 
 Thus, it is necessary to add the parameter $\alpha$ to make the value of the loss function greater than 0. CRL's convergence is better than the NLLR convergence.

Besides, when $\alpha$ is fixed to a constant greater than 0, $\beta$ changes from 0 to 1, the weight of competing class probabilities increases, and test errors have a tendency to decline; moreover, $\beta$ changes from 1 to 1.25, the weight of probabilities of competing classes exceed the weight of the correct class, and test errors increases. When $\beta$ is equal to 1, the test error rate is the lowest value. The updated sizes of backpropagation using CRL are the most suitable for CRL’s fastest convergence to local optimum. This indicates that when $\beta<1$, the update size of backpropagation is too large to converge to the local optimum. When $\beta>1$ the update size of backpropagation is too small, causing CRL to converge into the local optimum too slowly. When $\beta=1$, the ratio between competing classes and the correct class is the most suitable for CNN maintaining the highest confidence in the correct classification of images during backpropagation. In addition, when $\beta$ is fixed to a constant, test errors of $\alpha=1.5$ are lower than others.

To further compare the performance of CRL with NLLR and CEL, we conducted experiments using NLLR, CEL and CRL on CIFAR-10 with the experimental settings described in \cite{zhu2018negative}. The model in \cite{zhu2018negative} is defined as four convolutional layers followed by two fully-connected layers with max-pooling and dropout layers. The networks are trained for 100 epochs on CEL and 500 epochs on NLLR with a mini-batch size of 64. For comparison, we trained networks for 100 and 500 epochs on CRL ($\alpha=1.5, \beta=1$) separately. \textbf{Table \ref{nllr2}} shows test accuracies on CIFAR-10. In shallow neural networks, between 100 and 500 epochs, the difference of test accuracy using NLLR is larger than CEL and CRL. The test accuracy using NLLR in training 500 epochs is similar to the test accuracy using CRL in training 100 epochs. Besides, CRL and CE get closer to the best	results with fewer iterations than NLLR. This shows that NLLR convergence is obviously slower and less effective than CEL and CRL. Naturally, it is necessary to always use parameters that can increase the convergence speed of CRL.

Therefore, according to the ablation experiments in this section, in the following experiments, we set $\alpha=1.5$ and $\beta=1$, which are the most efficient hyperparameters.

\subsection{Experiments on different networks and different-depth networks}
\label{different model layers}
\begin{figure}[htbp]
	\begin{minipage}[b]{1.0\linewidth}
		\centering
		\centerline{\includegraphics[width=1\textwidth]{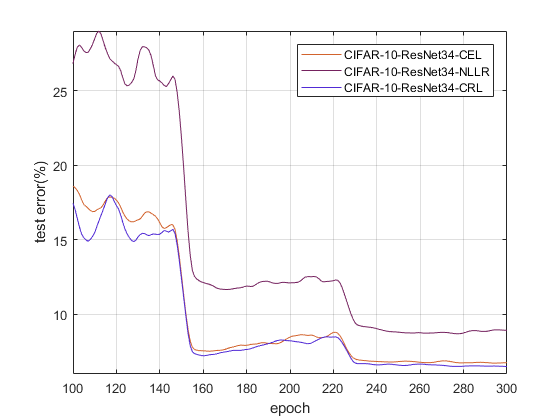}}
		%  \vspace{2.0cm}
		\centerline{(a) Smoothed test errors on CIFAR-10 by ResNet34}\medskip
	\end{minipage}
	\hfill
	\begin{minipage}[b]{1.0\linewidth}
		\centering
		\centerline{\includegraphics[width=1\textwidth]{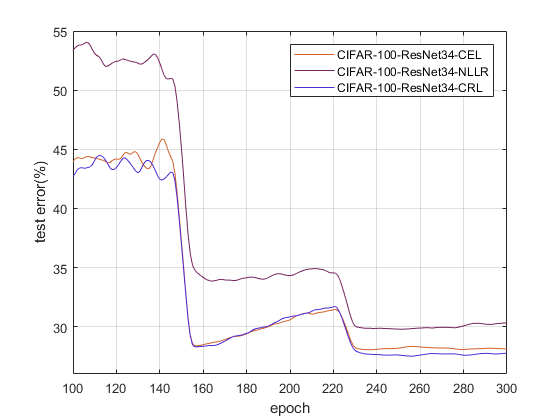}}
		%  \vspace{1.5cm}
		\centerline{(b) Smoothed test errors on CIFAR-100 by ResNet34}\medskip
	\end{minipage}	
	\caption{Smoothed test errors on CIFAR-10/100 by ResNet34}
	\label{testerror}	%
\end{figure}

In this section, we evaluate CRL on different networks and different-depth networks including ResNet34/50/101/164\cite{resnet}, VGG16/19\cite{vgg},  DenseNet\cite{huang2017densely}, and MobileNetV2\cite{Sandler_2018_CVPR}. \textbf{Table \ref{networkresult}} shows the test errors on CIFAR-10 and CIFAR-100. \textbf{Fig.\ref{testerror}} shows smoothed test errors on CIFAR-10/100 by ResNet34, corresponding to results in \textbf{Table \ref{networkresult}}. We can see that CRL outperforms CEL on different-depth and different kinds of models. These results demonstrate the effectiveness and robustness of CRL. No matter how many DCNN types or spin-offs are available, CRL can always achieve better results than its basic DCNN with the same number of layers. 
Besides, from \textbf{Fig.\ref{testerror}}, we can see that on CIFAR-10/100, CRL has a faster gradient convergence than CEL and NLLR from epoch 100- 150, when the learning rate is 0.1, because CRL is adaptive to the probability of a correct class in backpropagation.
 \textbf{Fig.\ref{testerror}} and \textbf{Table \ref{networkresult}} also show that NLLR performs worse than CRL and CEL on different networks and different-depth networks. From this it can be inferred that in the training procedure of DCNN, the gradient value of NLLR is not constantly positive, causing slow convergence, so NLLR is worse than CRL and CEL.

\begin{table}[htbp]
	\centering
	\caption{Test Errors (\%) on CIFAR-10/100 by different-depth and different kinds of networks}
	\begin{tabular}{|c|c|c|c|c|c|}
		\hline
		\multirow{2}{*}{\textbf{Dataset}} & \multicolumn{2}{c|}{\textbf{Network}} & \multirow{2}{*}{\textbf{NLLR}} & \multirow{2}{*}{\textbf{CEL}} & \multirow{2}{*}{\textbf{CRL}} \\ \cline{2-3}
		& \textbf{Type} & \textbf{\#Depth} &  &  &  \\ \hline \hline
		\multirow{9}{*}{\textbf{CIFAR-10}}  
		& \multirow{2}{*}{\textbf{VGG}\cite{vgg}} & 16 & 8.44 & 6.32 & \textbf{6.27} \\ 
		&  & 19 & 8.33 & 6.28 & \textbf{6.06 }\\ \cline{2-6} 
		& \multirow{4}{*}{\textbf{ResNet}\cite{resnet}} & 34 & 8.55 & 6.63 & \textbf{5.99} \\ 
		&  & 50 & 8.39 & 5.90 & \textbf{5.60} \\ 
		&  & 101 & 8.06 & 5.49 & \textbf{5.37} \\ 
		&  & 164 & 6.48 & 4.76 & \textbf{4.36 }\\  \cline{2-6}
		& \multirow{1}{*}{\textbf{DenseNet}\cite{huang2017densely}} & 40 & 6.46  & 5.32 & \textbf{5.26} \\  \cline{2-6}
		& \multirow{1}{*}{\textbf{MobileNetV2}\cite{Sandler_2018_CVPR}} &  54 & 6.83  & 6.01 & \textbf{5.53} \textbf{} \\ 
		\hline
		\multirow{6}{*}{\textbf{CIFAR-100}} 
		& \multirow{2}{*}{\textbf{VGG}\cite{vgg}} & 16 & 28.02 & 26.97 & \textbf{26.63} \\ 
		&  & 19 & 27.89 & 26.8 & \textbf{26.24} \\ \cline{2-6} 
		& \multirow{4}{*}{\textbf{ResNet}\cite{resnet}} & 34 & 29.67 & 27.87 & \textbf{27.34} \\ 
		&  & 50 & 27.26 & 25.33 & \textbf{25.23} \\ 
		&  & 101 & 26.77 & 24.27 & \textbf{23.34} \\ 
		&  & 164 & 25.02 & 22.26 & \textbf{21.94 }\\ \cline{2-6}
		& \multirow{1}{*}{\textbf{DenseNet}\cite{huang2017densely}} & 40 & 25.45   & 24.64  & \textbf{24.22} \\  \cline{2-6}
		& \multirow{1}{*}{\textbf{MobileNetV2}\cite{Sandler_2018_CVPR}} & 54 & 24.94 & 23.75  & \textbf{23.16} \\ 
		\hline
	\end{tabular}%
	\label{networkresult}%
\end{table}

\begin{table}[]
	\centering
	\caption{Test Errors (\%) on MNIST by simple classifiers}
	\begin{tabular}{|c|c|c|c|c|}
		\hline
		\textbf{Model}                  & \textbf{Loss}       & \textbf{ERROR}  & \textbf{F1-score} \\ \hline \hline
		
		\textbf{Logistic Regression}    & \textbf{CEL}        & 9.20   & 90.71
		\\ \hline
		\textbf{Logistic Regression}    & \textbf{CRL}        & \textbf{8.60} & \textbf{91.3}
		\\ \hline
		
		\textbf{Multilayer Perceptron}  & \textbf{CEL}        & 9.10  & 90.77   \\ \hline
		\textbf{Multilayer Perceptron}  & \textbf{CRL}        & 8.85  & 91.03 \\ \hline
	\end{tabular}
	\label{tranditopnalresult}%
\end{table}

\begin{table}[]
	\centering
	\caption{Compares of the classificartion performances on CIFAR-100 by MobileNetV2. Each method runs 10 times.}
	\begin{tabular}{|c|c|c|c|}
		\hline
		\textbf{Methods} & \textbf{Mean}  & \textbf{Std.} & \textbf{p-Value} \\ \hline
		\textbf{CEL}    & 76.46 & 0.28  &    \multirow{2}{*}{0.00136}         \\ \cline{1-3}
		\textbf{CRL}     & \textbf{76.79} & \textbf{0.20}    &    \\ \hline  
	\end{tabular}
	\label{meanresults}%
\end{table}

In order to more effectively prove the effectiveness of our method, we compare our proposed loss function with the traditional loss functions used in Logistic Regression and Multilayer Perceptron. \textbf{Table \ref{tranditopnalresult}} shows the test errors and F1-scores on MNIST. We can see that CRL outperforms CEL on Logistic Regression and Multilayer Perceptron.

Moreover, to demonstrate that our  CE loss is not due to chance, we run CE-loss and our-loss on CIFAR-100 by MobileNetV2 10 times each. The means and standard deviations of the accuracies are shown in \textbf{Table \ref{meanresults}}. The means and standard deviations of the accuracies are 76.46 and 0.28 for CE Loss, and 76.79 and 0.20 for our loss.
We conduct paired Student's t-tests between CE-loss and our-loss on CIFAR-100 by MobileNetV2.
The p-value of the paired Student's t-test for CEL and CRL is 0.00136, which is less than 0.005(significance level). Therefore, we reject the null hypothesis that CEL and CRL have the same mean accuracy. And our loss is significantly different with the CE loss.

\subsection{Generalization on some image classification datasets}
\label{Generalization on some image classification datasets}
\begin{figure*}[htbp]
	\centering
	\subfigure[Good results of using our CRL and CEL.]{
		\begin{minipage}[]{0.5\linewidth}
			\centering
			\centerline{\includegraphics[width=2\textwidth]{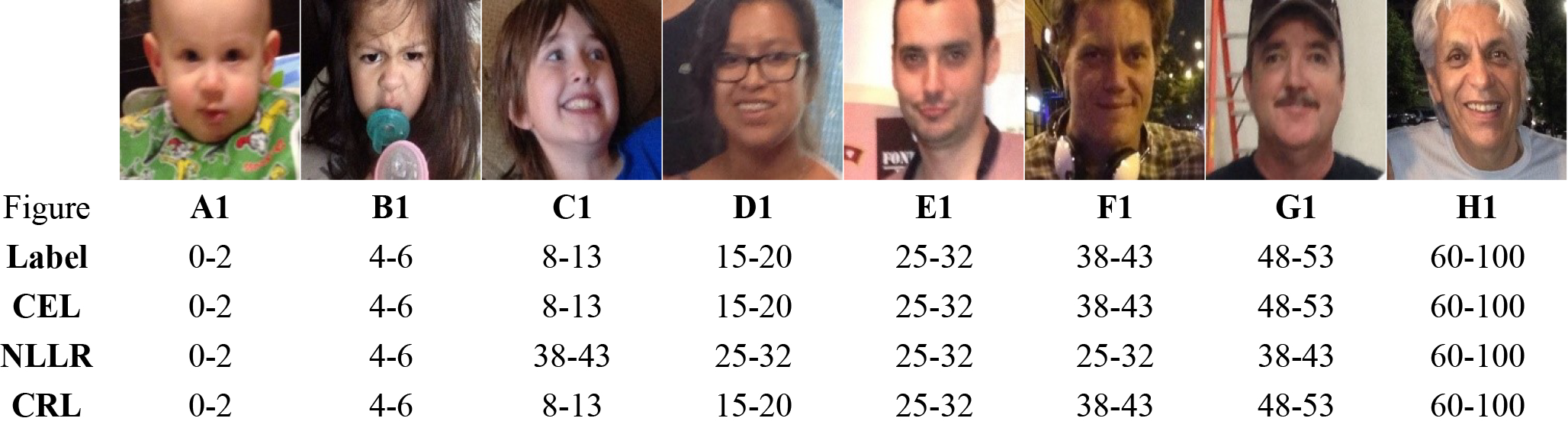}}
		\end{minipage}%
	}%
	
	\subfigure[Good results using our CRL but bad  results using CEL and NLLR]{
		\begin{minipage}[]{0.5\linewidth}
			\centering
			\centerline{\includegraphics[width=2\textwidth]{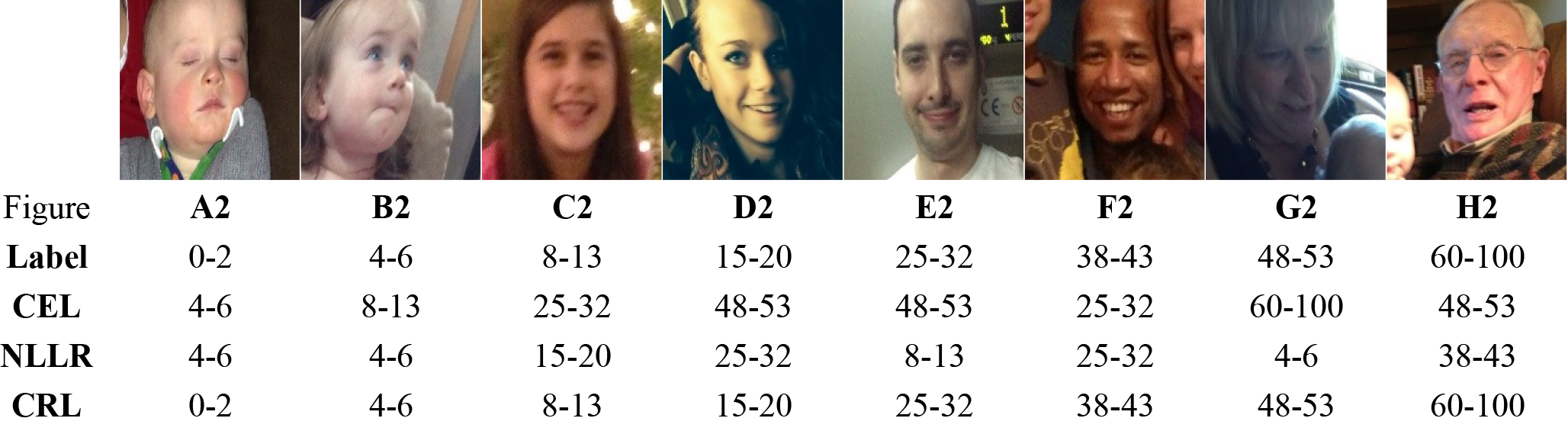}}
		\end{minipage}%
	}%
	
	\centering
	\caption{Examples of age group classification results by ResNet34 using CEL and CRL. The top row shows good results of using our CRL and CEL. The below row shows good results using our CRL but bad results using CEL and NLLR. The table below each image shows the labeled class and the classification results using our CRL and CEL.}
	\label{examples}
	
\end{figure*}
\begin{figure*}[htbp]
	\subfigure[The softmax outputs of  image A1-H1 using CEL]{
		\begin{minipage}[]{0.33\linewidth}
			\centering
			\includegraphics [width=1\textwidth] {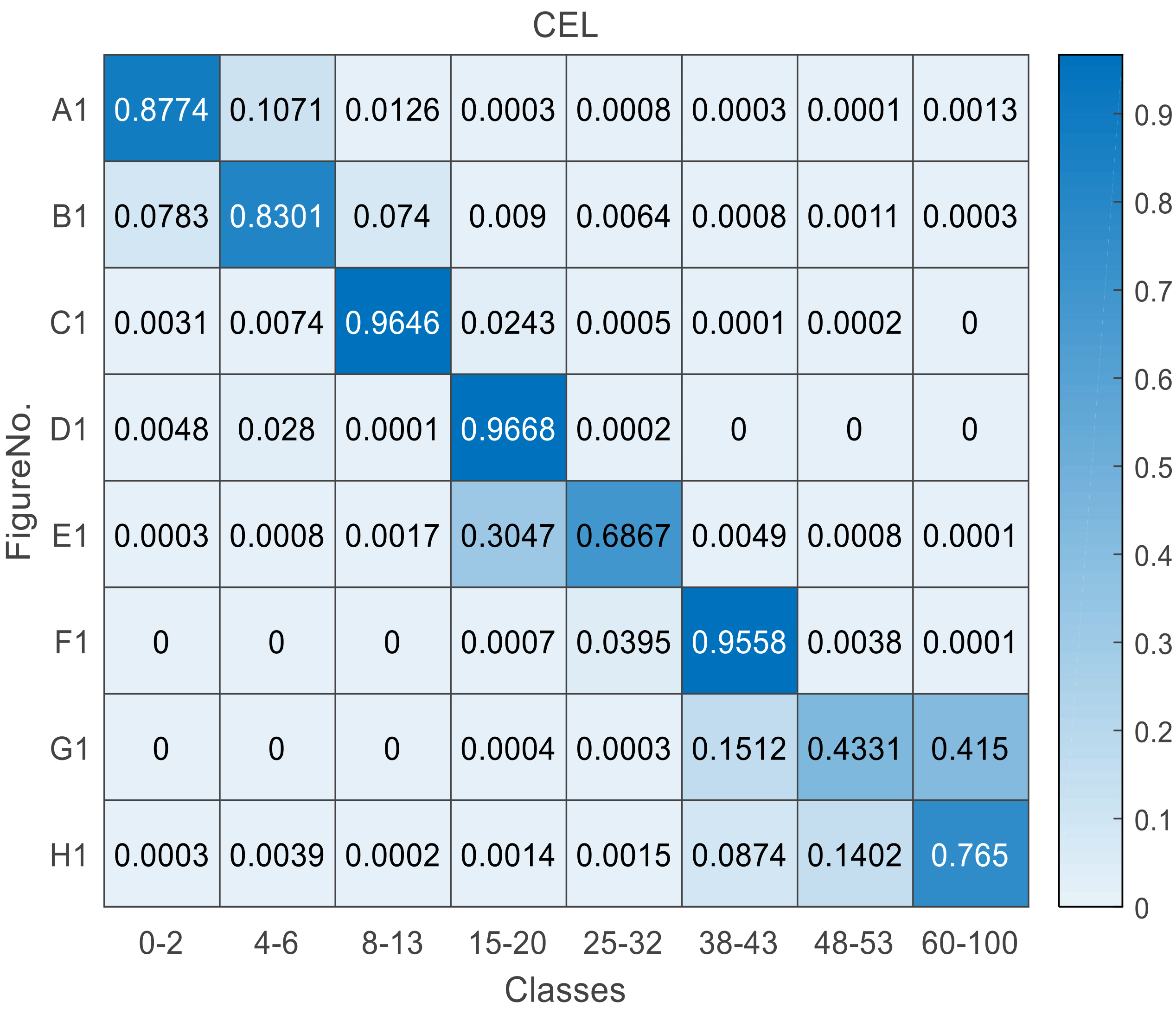}	
		\end{minipage}% 
	}
	\subfigure[The softmax outputs of  image A1-H1 using CRL]{
		\begin{minipage}[]{0.33\linewidth}
			\centering
			\includegraphics [width=1\textwidth] {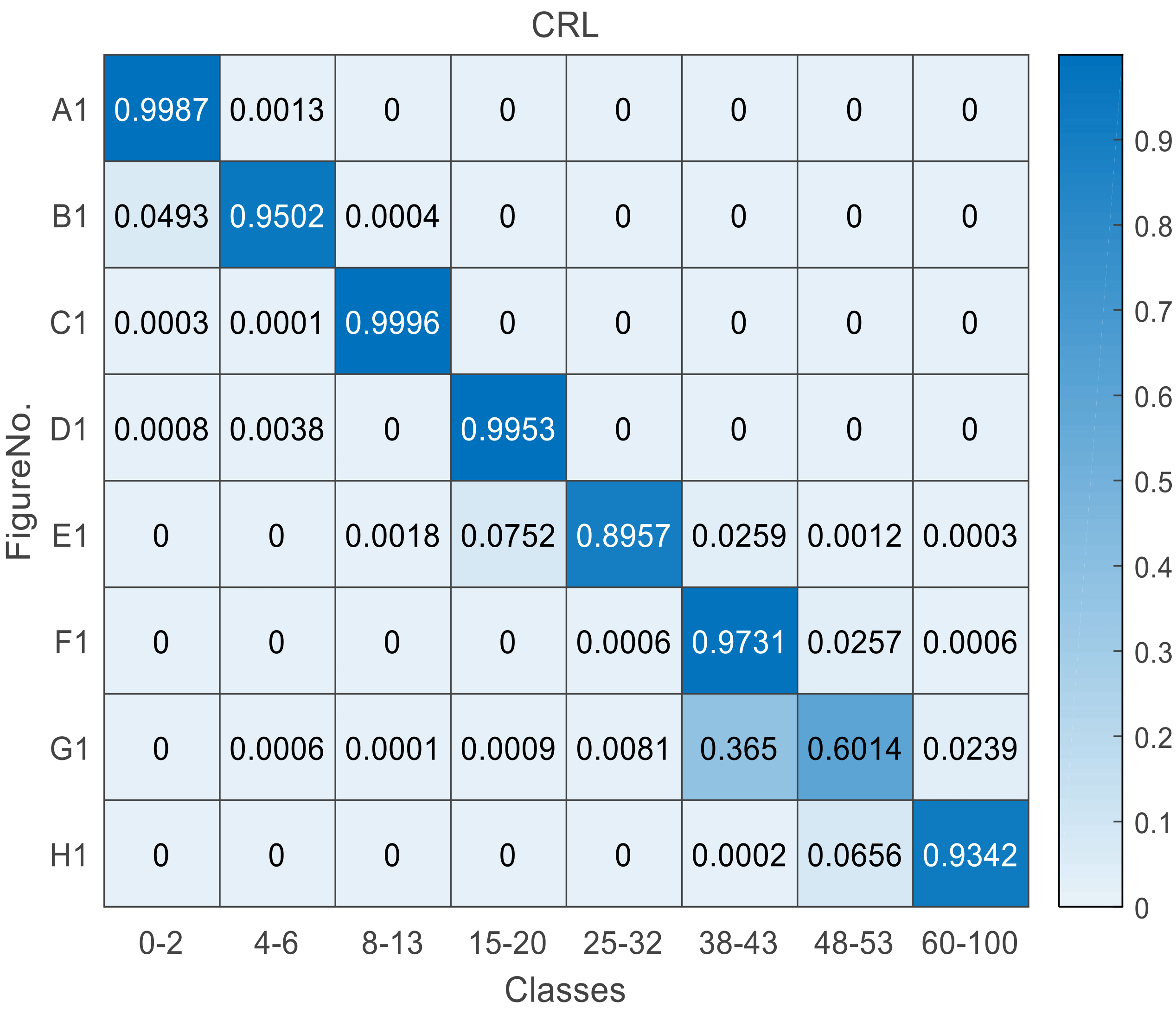}	
		\end{minipage}% 
	}
	\subfigure[The softmax outputs of  image A1-H1 using NLLR]{
		\begin{minipage}[]{0.33\linewidth}
			\centering
			\includegraphics [width=1\textwidth] {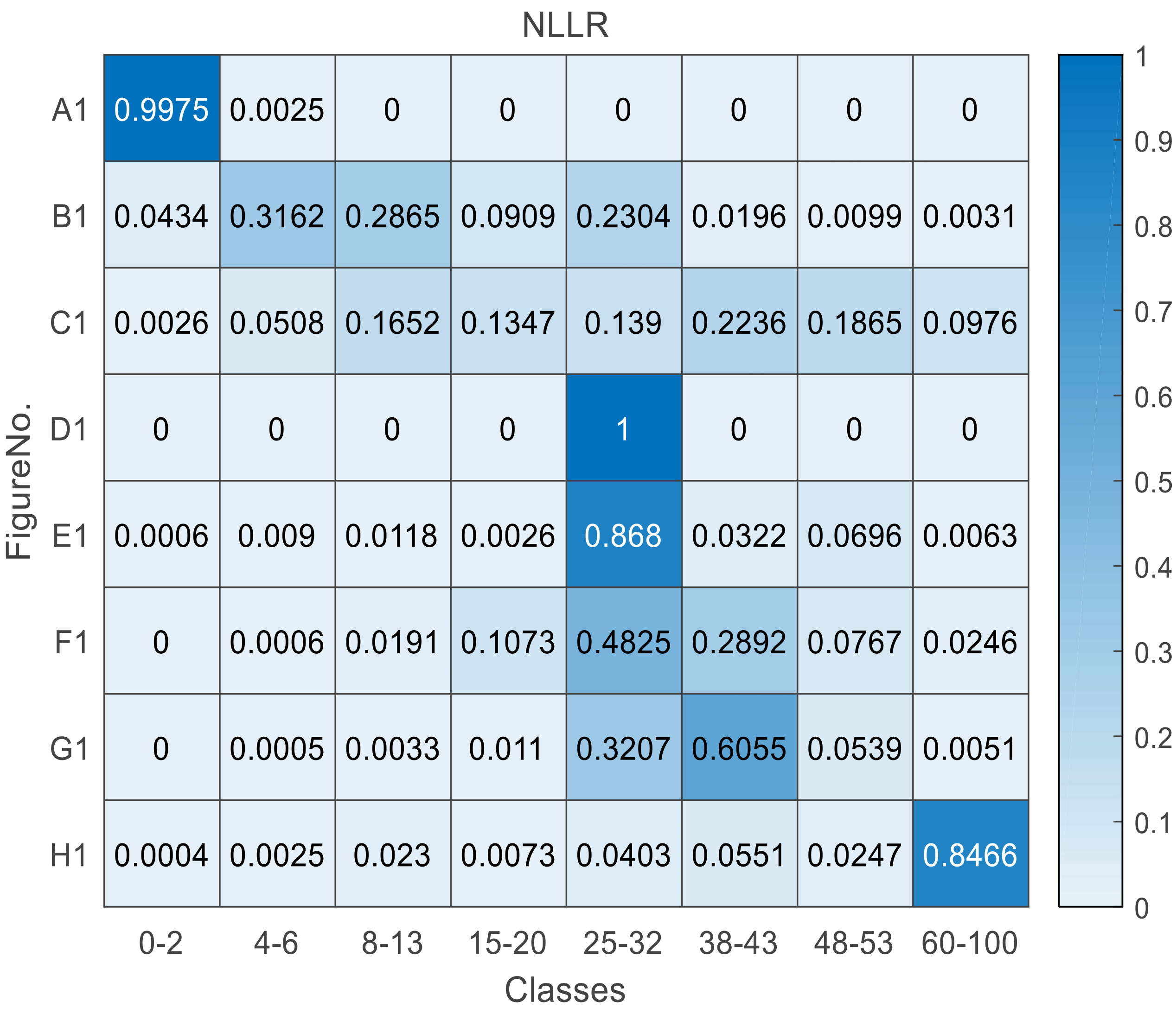}	
		\end{minipage}% 
	}
	
	\subfigure[The softmax outputs of  image A2-H2 using CEL]{
		\begin{minipage}[]{0.33\linewidth}
			\centering
			\includegraphics [width=1\textwidth] {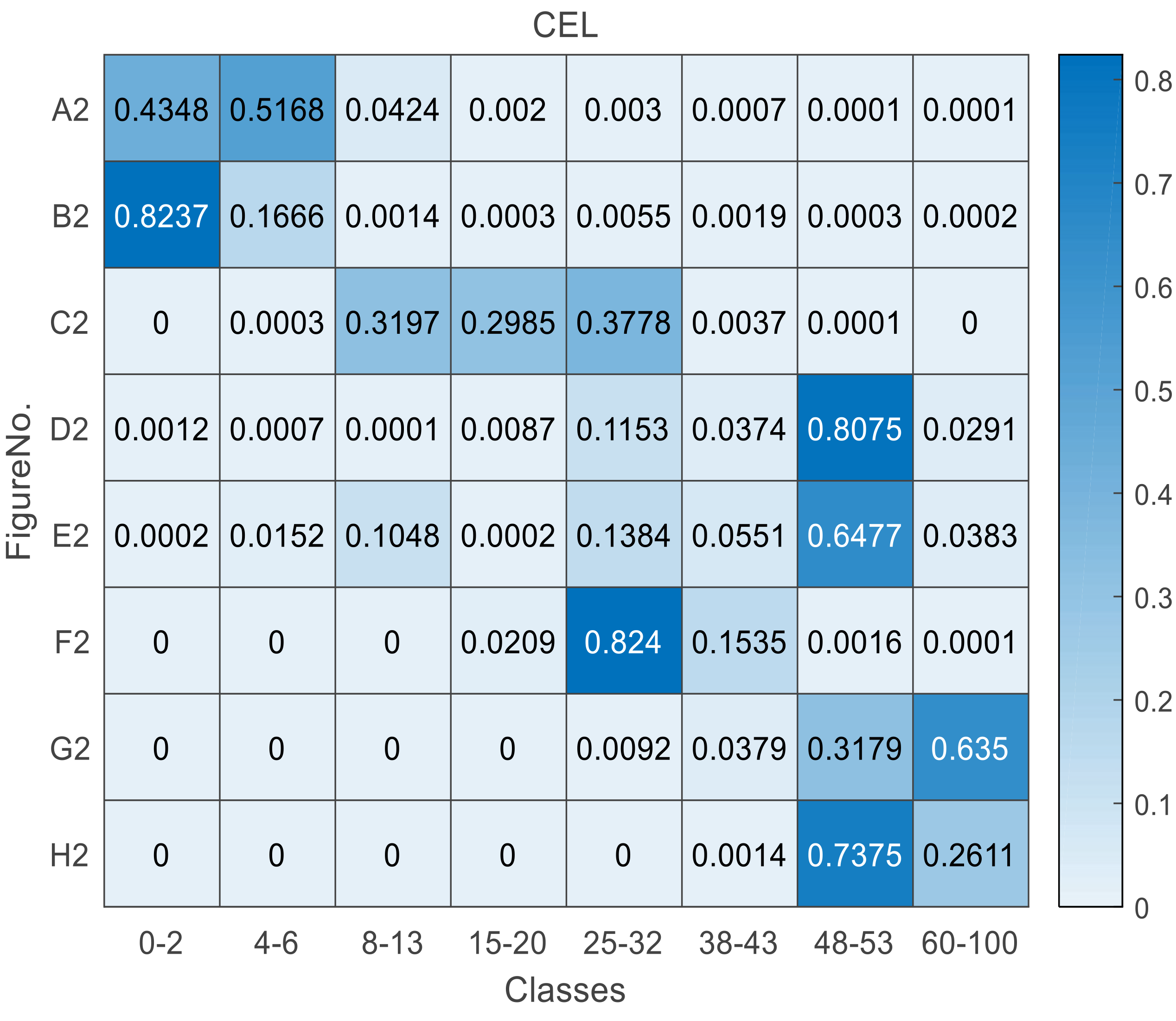}
		\end{minipage}% 
	}
	\subfigure[The softmax outputs of  image A2-H2 using CRL]{
		\begin{minipage}[]{0.33\linewidth}
			\centering
			\includegraphics [width=1\textwidth] {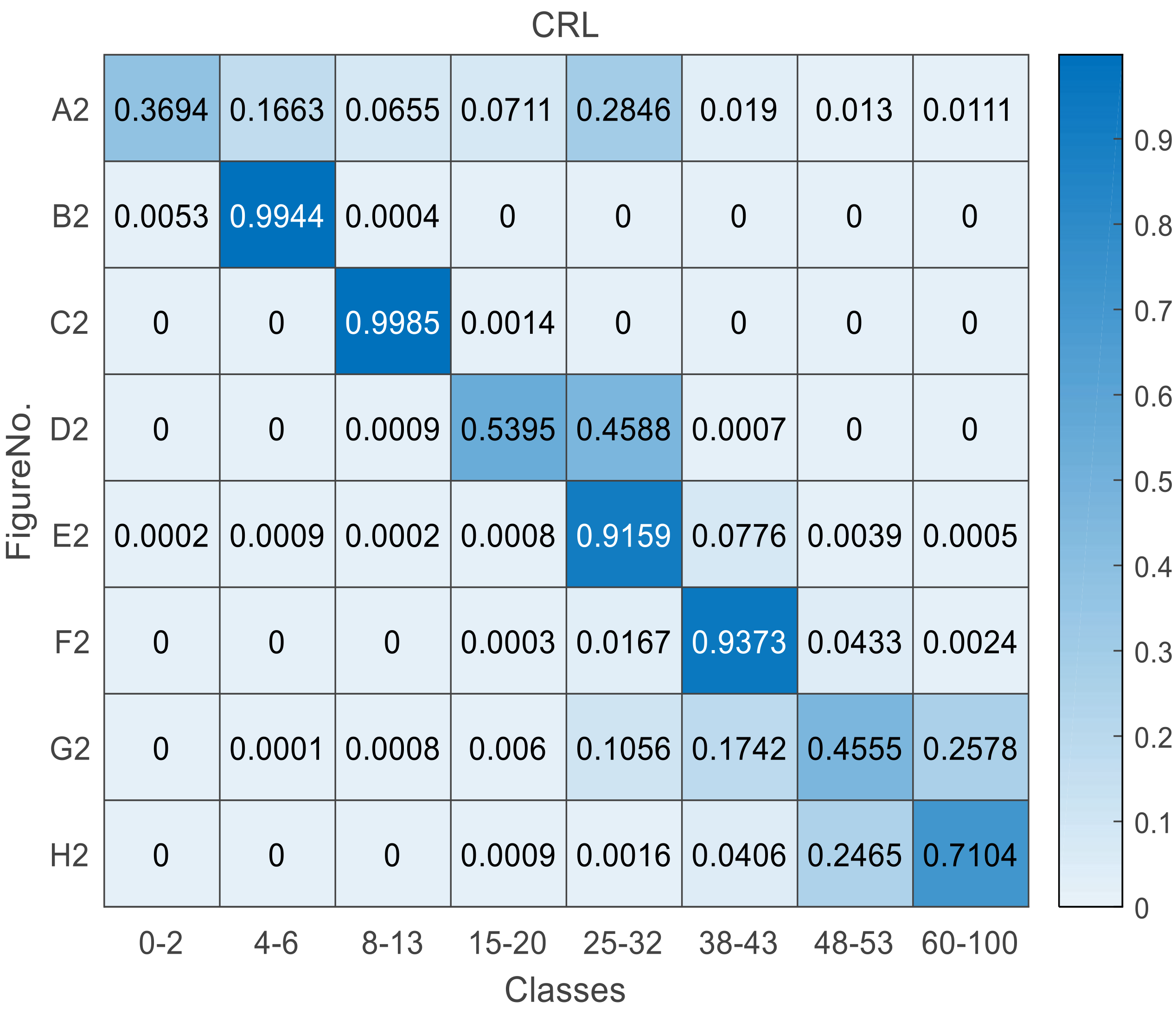}
		\end{minipage}% 
	}
	\subfigure[The softmax outputs of  image A2-H2 using NLLR]{
		\begin{minipage}[]{0.33\linewidth}
			\centering
			\includegraphics [width=1\textwidth] {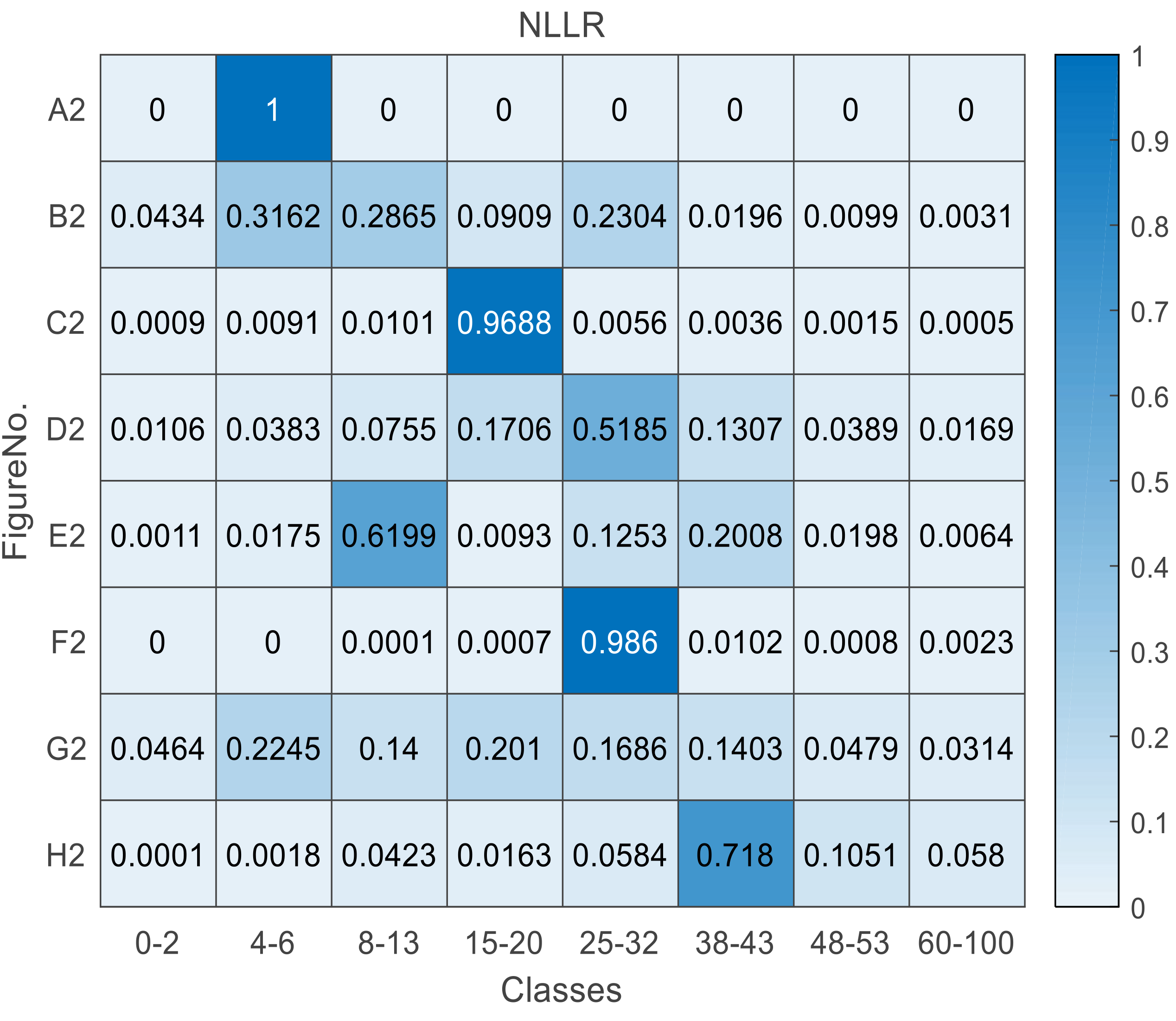}
		\end{minipage}% 
	}
	\caption{The softmax output heatmap of example images corresponds to \textbf{Fig.\ref{examples}}. The horizontal axis of each heatmap represents the prediction class. The vertical axis represents the example image's number corresponding to \textbf{Fig.\ref{examples}}. Each row in heatmap is the probability of ten classes for one image. Subfigure (a), (b) and (c)  correspond to (a) in \textbf{Fig.\ref{examples}},  Subfigure (d), (e) and (f) correspond to (b) in \textbf{Fig.\ref{examples}}.}
	\label{heatmap}
\end{figure*}

To experimentalize the generalization of our proposed CRL in different image classification tasks, we performed a series of experiments on general image classification datasets, fine-grained image classification datasets, and a face-age estimation dataset.

\subsubsection{General image classification tasks}
\label{general image classification tasks}

A typical general image classification problem is the task of assigning an input image, which is one label away from a fixed set of categories, which is a research focus of computer vision. We tested our proposed loss on CIFAR10/100, SVHN and ImageNet. CIFAR10/100 and SVHN are small-scale image classification datasets. ImageNet is a challenging large-scale dataset.

\textbf{Table \ref{differentloss}} Table IV compares CRL and other popular loss functions by the ResNet164 model on CIFAR-10/100 datasets, including cross-entropy loss (CEL), negative log likelihood ratio loss (NLLR) \cite{zhu2018negative}, focal loss (FL)\cite{focalloss}, large-margin loss (L-Softmax)\cite{largemarginloss}, Angular Softmax (A-Softmax)\cite{A-Softmaxloss}, Dual cross-entropy loss (DualCE)\cite{DualCE} and taming cross-entropy loss (TCE)\cite{tamingloss}. We don't directly use the results in the original papers, but reran the loss functions on the ResNet164 model, because the DCNN with deep layers has a better generalization, which makes the comparison of loss functions more obvious. Our CRL outperformed the other loss functions, which obviously indicates that CRL is effective. Through experiments, we argue that our CRL can improve the discrimination ability by computing the competing ratio between correct class and competing incorrect classes.

\begin{table}[htbp]
	\centering
	\caption{Test errors (\%) on CIFAR-10/100 using different loss functions by ResNet164}
	\begin{tabular}{|c|c|c|}
		\hline
		\textbf{Loss Function}&\textbf{CIFAR-10}&\textbf{CIFAR-100}\\\hline\hline
		\textbf{CEL}\cite{resnet}&5.27&22.26\\\hline
		\textbf{NLLR}\cite{zhu2018negative}&6.48&25.02\\\hline
		\textbf{FL}\cite{focalloss}&4.62&22.13\\\hline
		\textbf{L-Softmax}\cite{largemarginloss}&4.71&22.21\\\hline
		\textbf{A-Softmax}\cite{A-Softmaxloss}&4.64&22.12\\\hline
		\textbf{DualCE}\cite{DualCE}&4.58&22.43\\\hline
		\textbf{TCE}\cite{tamingloss}&5.46&23.93\\\hline
		\textbf{CRL}&\textbf{4.36}&\textbf{21.94}\\\hline
	\end{tabular}
	\label{differentloss}	
\end{table}

\textbf{Table \ref{generalization}} shows classification error rates (\%) on general image classification datasets using CRL, CEL, and NLLR. On CIFAR10/100 and SVHN datasets, we used ResNet164 as the backbone network. Limited to computing power, we only used ResNet34 as the backbone network on the ImageNet dataset. Generally, we can see that NLLR has worse generalization than CRL and CEL in the datasets above. Thus, improving NLLR is necessary. On the SVHN dataset, the classification error rate of CRL is 1.88\%, which is about 7.3\% lower than CEL. Then, experimental result showed that the proposed loss achieves competitive performance on the SVHN dataset. 
In addition, \textbf{Table \ref{generalization}} also shows the Top-1 and Top-5 errors associated with using CRL and CEL on ImageNet, which is currently the largest large-scale image classification dataset. Compared with CEL, our CRL absolutely outperforms by 0.15\% on the Top-1 error, 0.26\% on Top-5 error. Because the amount of ImageNet is large enough, both CRL and CEL can fit the ImageNet well. Even though the improvement of CRL is not high, it verifies the effectiveness of CRL on large-scale image datasets.

Moreover, to demonstrate that our  CE loss is not due to chance, we run CE-loss , NLLR-loss and our-loss on CIFAR-10 by ResNet164 10 times each. The means and variances of errors are shown in \textbf{Table \ref{meanresults-1}}. The means and variances of the errors are 6.61 and 0.127 for NLLR Loss, and 5.38 and 0.071 for CE loss. And the mean and variance of our loss is 4.48 and 0.073. We can conclude that the superiority of the proposed method is not caused by the parameter fluctuation.
\begin{table}[htbp]
	\centering
	\caption{Classification error rates(\%) on general image classification datasets using CRL, CEL and NLLR. In the ImageNet line, the content outside the brackets represents Top-1 error, the content in parentheses represents Top-5 error\cite{imageNet}.}
	\begin{tabular}{|c|c|c|c|}
		\hline 
		\textbf{Dataset} & \textbf{NLLR} & \textbf{CEL}& \textbf{CRL} \\ 
		\hline \hline
		\textbf{CIFAR10} & 6.48 & 5.27 & \textbf{4.36} \\ 
		
		\textbf{CIFAR100} & 25.02 & 22.26 & \textbf{21.94} \\ 
		
		\textbf{SVHN} & 2.38 & 2.04 & \textbf{1.88} \\ 
		 
		\textbf{ImageNet} &--- & 27.09(8.95) & \textbf{26.94(8.69)}  \\ 
		\hline 
	\end{tabular}
	\label{generalization}
\end{table}

\begin{table}[]
	\centering
	\caption{Compares of the classificartion performances on CIFAR-10 by  ResNet164. Each method runs 10 times.}
	\begin{tabular}{|c|c|c|c|}
		\hline
		\textbf{Methods} & \textbf{Mean error}  & \textbf{Var.} \\ \hline \hline
		\textbf{NLLR}    & 6.61 & 0.127     \\ 
		\textbf{CEL}    & 5.38 & 0.071    \\ 
		\textbf{CRL}     & \textbf{4.48} & \textbf{0.073}        \\ \hline  
	\end{tabular}
	\label{meanresults-1}%
\end{table}

\subsubsection{Fine-grained image classification tasks}
\label{fine-grained image classification tasks}
\begin{table}[]
	\centering
	\caption{Comparison results on CUB200-2011 and Standford Cars datasets. Base Model means the backbone network used in the method. The Contents in the parentheses mean the rerun results.}

	\begin{tabular}{|c|c|c|c|c|}
		\hline
		\multirow{2}{*}{\textbf{Method}} & \multirow{2}{*}{\textbf{Base Model}} &  \multicolumn{2}{c|}{\textbf{Accuracy(\%)}} \\\cline{3-4}
		&  & \textbf{CUB} & \textbf{STCAR} \\ \hline \hline
		\textbf{B-CNN}\cite{BCNN}& VGGNet  &84.1(84.06)& 91.3(91.27)\\
		\textbf{RA-CNN}\cite{RACNN}&VGG19&85.3&92.5\\
		\textbf{OPAM}\cite{OPAM}&VGG16&85.8&92.2\\
		\textbf{Kernel-Pooling}\cite{Kernalpooling}&VGG16&86.2&92.4\\
		\textbf{Kernel-Pooling}\cite{Kernalpooling}&ResNet50&84.7&91.1\\
		\textbf{MA-CNN}\cite{MACNN}&VGG-19&86.5&92.8\\
		\textbf{DFL}\cite{DFLCNN}& ResNet50  & 87.4(87.14) & 93.8(93.79) \\
		\textbf{DCL}\cite{DCL}& ResNet50  & 87.8(87.69) & 94.5(94.39) \\\hline
		\textbf{B-CNN+CRL}& VGGNet & 84.27 &  91.44\\		
		\textbf{DFL+CRL}& ResNet50  & 87.52 & 93.95 \\ 
		\textbf{DCL+CRL}& ResNet50 & \textbf{87.84} & \textbf{94.60} \\ \hline
		
	\end{tabular}
	\label{fine-grained result}
\end{table}

We compared fine-grained image classification with general image classification and found the differences and difficulties of fine-grained image classification to be the more subordinate categories of the super-category, which the fine-grained images belong to. To verify the effectiveness of our proposed loss on fine-grained image classification tasks, we tested our proposed CRL and CEL performance on CUB200-2011 and Standford Car datasets. CUB200-2011. We found Standford Cars to be the most challenging datasets, which aim to distinguish subordinate-level bird species and car categories. For better comparison, we used ResNet50 and VGGNet pretrained on ImageNet as the backbone network, which are the most popular networks on the fine-grained image classification task. Besides, we reran the CE and CRL image reclassification datasets using the experimental setting described in \cite{BCNN,DFLCNN,DCL}. The comparison results on CUB200-2011 and Standford Cars datasets are displayed in \textbf{Table \ref{fine-grained result}}. The results showed that our proposed approach can be applied to various methods. The accuracy of CRL is better than that of repeated experiments or that of the original paper. Besides, \textbf{DCL+CRL} achieves better results than the state-of-the-art methods on CUB200-2011. It shows that, for fine-grained image classification, our CRL can better discriminate fine-grained class from the competing classes, improving the accuracy of CEL and model discriminability.

\subsubsection{Face age estimation task}
\label{Face age estimation task}
\begin{table}[]
	\centering
	\caption{The exact classification accuracy and with-one-category-off accuracy\cite{fine-grainedlstm} using CRL, CEL and NLLR by different depth networks including ResNet34/101/152. "Pre-" means the model is pretrained on ImageNet.}
	
		\begin{tabular}{|c|c|c|c|}
			\hline
			\multicolumn{2}{|c|}{\textbf{Method}} &\textbf{Age Exact Acc(\%)} & \textbf{Age 1-off Acc(\%)} \\
			\hline\hline
			\multicolumn{1}{|c|}{\multirow{3}{*}{\textbf{ResNet34}}} & NLLR  & 55.27 & 89.30  \\
			\cline{2-4}          & CEL   & 56.32 & 90.99  \\
			\cline{2-4}          & CRL   & 57.05 & 91.26  \\
			\hline
			\multicolumn{1}{|c|}{\multirow{2}{*}{\textbf{Pre-ResNet34}}} & CEL   & 61.09 & 92.06  \\
			\cline{2-4}          & CRL   & 61.52 & 92.65  \\
			\hline
			\multicolumn{1}{|c|}{\multirow{2}{*}{\textbf{Pre-ResNet101}}} & CEL   & 62.21 & 92.18  \\
			\cline{2-4}          & CRL   & 62.73 & 92.79  \\
			\hline
			\multicolumn{1}{|c|}{\multirow{2}{*}{\textbf{Pre-ResNet152}}} & CEL   & 62.85 & 92.87  \\
			\cline{2-4}          & CRL   & 63.26 & 93.25  \\
			\hline
		\end{tabular}%

	\label{adience}
\end{table}

 As an emerging biometric recognition technology, age estimation technology based on face image is an important research subject in the field of computer vision. The Adience dataset is a popular age estimation dataset, which is difficult to classify due to the similarity between adjacent age groups. 

In order to promoting performance of face age estimation, the models are usually pretrained on ImageNet and finetuned on age datasets. So, we tested our CRL and CEL on the Adience dataset while using different depth networks including ResNet34/101/152 pretrained on ImageNet. For comparison, we tested CRL, CEL and NLLR by ResNet34 without any pretraining on ImageNet. \textbf{Table \ref{adience}} shows the exact classification accuracy with-one-category-off accuracy\cite{fine-grainedlstm} using CRL and CEL by ResNet34/101/152. CRL achieves better performance than CEL on different-depth networks. CRL is more robust. \textbf{Fig.\ref{examples}} shows examples of age group classification results by ResNet34 using CEL, NLLR and CRL on the Adience dataset. The top row shows good results after using our CRL and CEL. The row below shows good results using our CRL but bad results using CEL. We can observe that the proposed loss is more robust to most of the common facial appearance variations than CEL, such as multi-face in the image (F2, G2), illumination (D2), filming angle (H2) and so on. We can also see that NLLR gets very poor performance on predicting ages than CEL or CRL.  \textbf{Fig.\ref{heatmap}} shows the softmax output heatmap of example images corresponding to \textbf{Fig.\ref{examples}}. We can see that our proposed loss places the softmax output on the diagonal line of the heatmap, which shows CRL making the prediction probability of the correct class as high as possible, which weakens the prediction probabilities of the competing incorrect classes. For example, observing the softmax output of the image G1 using CRL and CEL, we can conclude that when the prediction probabilities of some adjacent classes including the correct class, our CRL can make the prediction probability of the correct class and finish with a higher value than other adjacent classes based on selection of the correct class. Observing the softmax output of image C2 using CRL and CEL, our CRL can concentrate on the correct class; however, when CEL fails to classify correctly, the results demonstrated the consistent improvements of our CRL over the CEL in the hard face age estimation tasks.

\section{Conclusions}
\label{sec:conclusion}
In this study, we proposed the competing ratio loss function (CRL) which discriminates the correct class probability from the competing incorrect ones. The proposed loss calculates the ratio between the correct class probability and competing incorrect classes. This ratio strengthens the posterior probability of the correct class and weakens the posterior probabilities of incorrect classes. The results of a series of experiments showed that compared with CEL and NLLR, our proposed competing loss function (CRL) (1) has a better discrimination ability to detect the correct class and the competing incorrect classes; (2) CRL is constantly positive, which ensures CRL’s ability to achieve the optimization as a minimization problem in training DCNN, which means CRL’s convergence is better than NLLR; (3) CRL has strong applicability to different types and depths of DCNNs; (4) CRL has better performance on different image classification tasks, including general image classification, fine-grained image classification, and difficult face age estimation tasks.

\section*{Acknowledgments}

The authors gratefully acknowledge the support of NVIDIA Corporation with the donation of the GPU used for this research.

\ifCLASSOPTIONcaptionsoff
  \newpage
\fi
\bibliographystyle{IEEEtran}
\bibliography{refen}
\begin{IEEEbiography}[{\includegraphics[width=1in,height=1.25in,clip,keepaspectratio]{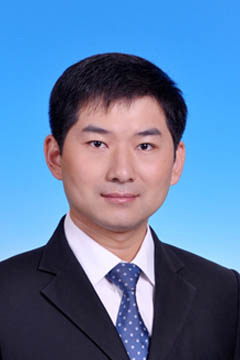}}]{Ke Zhang}
	received the M.E. degree in signal and information processing from North China Electric Power University, Baoding, China, in 2006, and the Ph.D. degree in signal and information processing from the Beijing University of Posts and Telecommunications, Beijing, China, in 2012. He finished his Post Doc in computer vision from the University of Missouri, Columbia, MO, USA, in 2016. He is currently an Associate Professor with North China Electric Power University. His research interests include computer vision, deep learning, machine learning, robot navigation, natural language processing, and spatial relation description. 
\end{IEEEbiography}
\vspace{-5pt}
\begin{IEEEbiography}[{\includegraphics[width=1in,height=1.25in,clip,keepaspectratio]{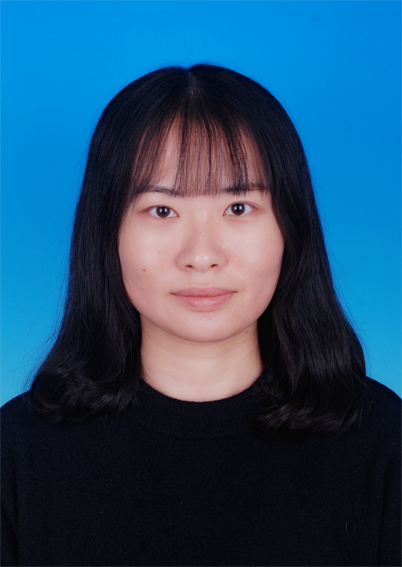}}]{Yurong Guo}
	received the M.E. degree in  communication and information engineering from North China Electric Power University, Baoding, China, in 2020, where she is currently a Ph.D. student in Beijing University of Posts and Telecommunications (BUPT). Her research interests include computer vision and deep learning. 
\end{IEEEbiography}
\vspace{-5pt}

\begin{IEEEbiography}[{\includegraphics[width=1in,height=1.25in,clip,keepaspectratio]{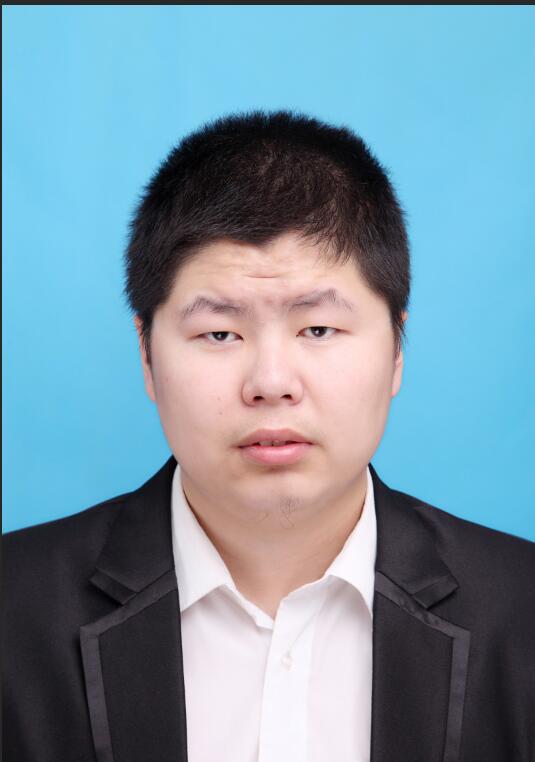}}]{Xinsheng Wang}
	received the B.S. degree in electronic information science and technology from North China Electric Power University, Baoding, China, in 2017, and he received the M.S. degree in communication and information engineering from North China Electric Power University, Baoding, China, in 2020. 
\end{IEEEbiography}
\vspace{-5pt}

\begin{IEEEbiography}[{\includegraphics[width=1in,height=1.25in,clip,keepaspectratio]{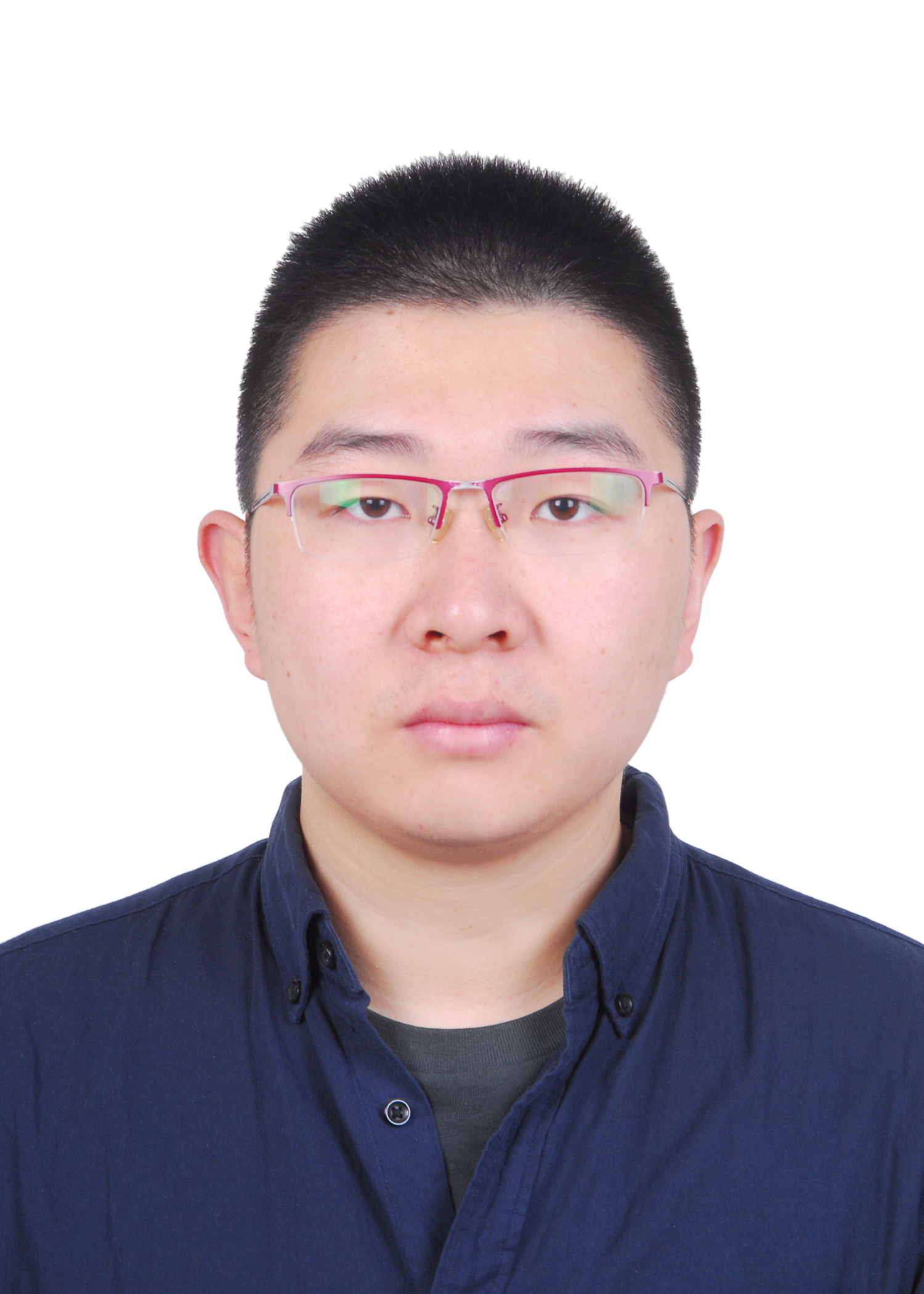}}]{Dongliang Chang}
	received his M.E. degree in Internet of things engineering from Lanzhou University of Technology, China, in 2019. He is currently a Ph.D. student in Beijing University of Posts and Telecommunications (BUPT). His research interests include machine learning and computer vision.
\end{IEEEbiography}
\vspace{-5pt}
\begin{IEEEbiography}[{\includegraphics[width=1in,height=1.25in,clip,keepaspectratio]{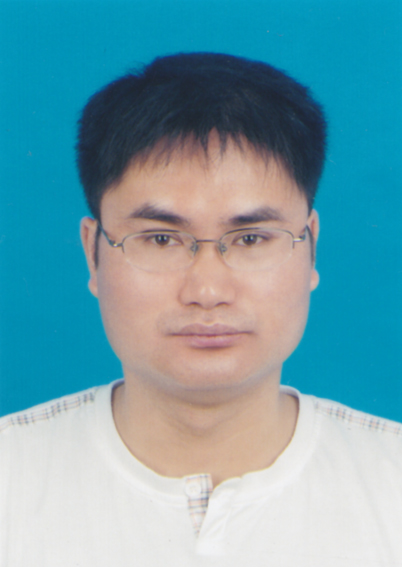}}]{Zhenbing Zhao}
	was born in Suqian, Jiangsu, China, in 1979. He received the B.S., M.S., and Ph.D. degrees from North China Electric Power University, Baoding, in 2002, 2005, and 2009, respectively. He is currently an Associate Professor with the School of Electrical and Electronic Engineering, North China Electric Power University. His research interests include machine learning, image processing, and the intelligent detection of electrical equipment. 
\end{IEEEbiography}
\vspace{-5pt}
\begin{IEEEbiography}[{\includegraphics[width=1in,height=1.25in,clip,keepaspectratio]{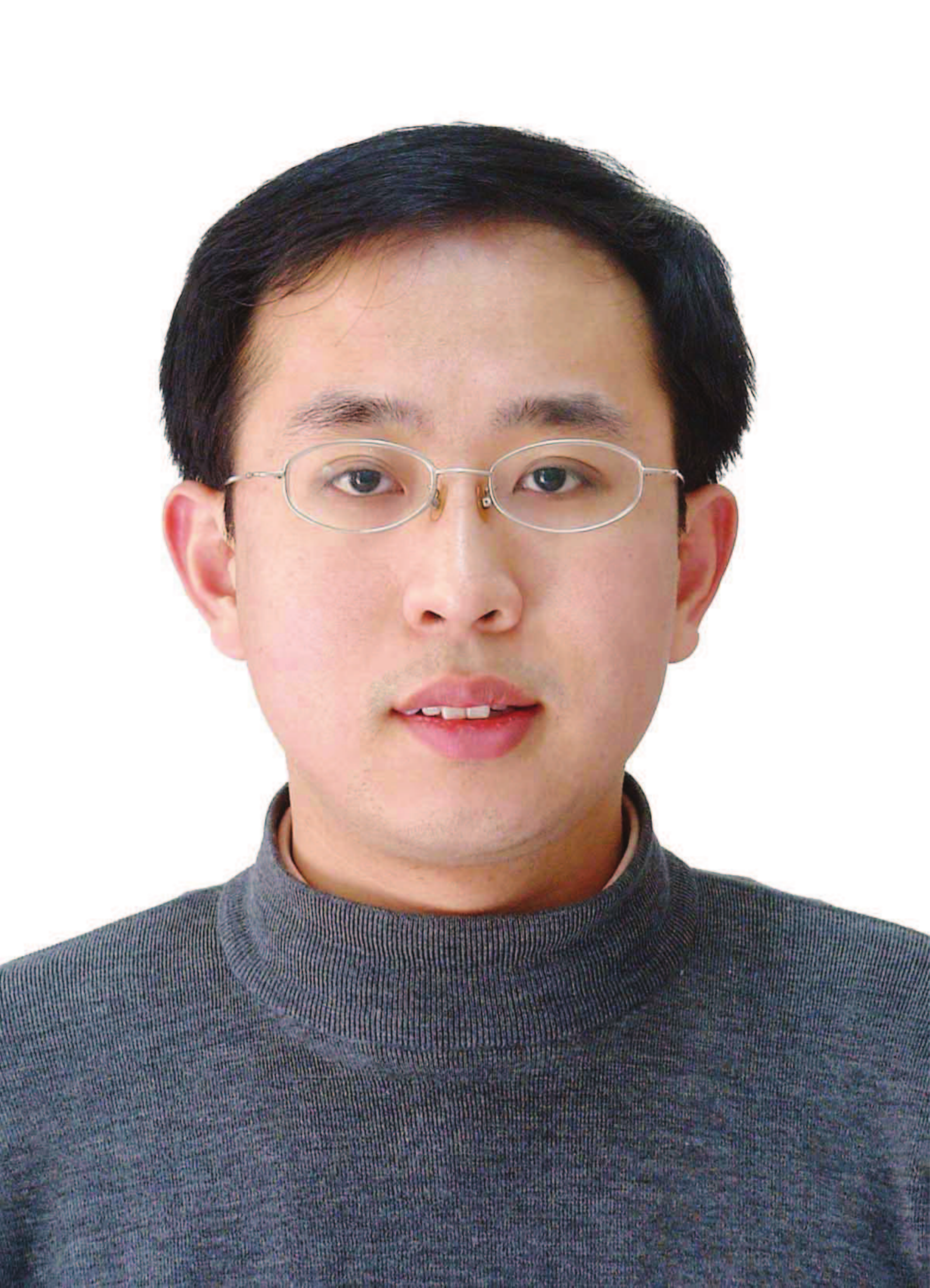}}]{Zhanyu Ma} is currently a Professor at Beijing University of Posts and Telecommunications, since $2019$. He received the Ph.D. degree in electrical engineering from the KTH Royal Institute of Technology, Sweden, in $2011$. From $2012$ to $2013$, he was a Postdoctoral Research Fellow with the School of Electrical Engineering, KTH Royal Institute of Technology. He has been an Associate Professor with the Beijing University of Posts and Telecommunications, Beijing, China, from $2014$ to $2019$. His research interests include pattern recognition and machine learning fundamentals with a focus on applications in computer vision, multimedia signal processing, data mining. He is a Senior Member of IEEE.
\end{IEEEbiography}
\vspace{-5pt}
\begin{IEEEbiography}[{\includegraphics[width=1in,height=1.25in,clip,keepaspectratio]{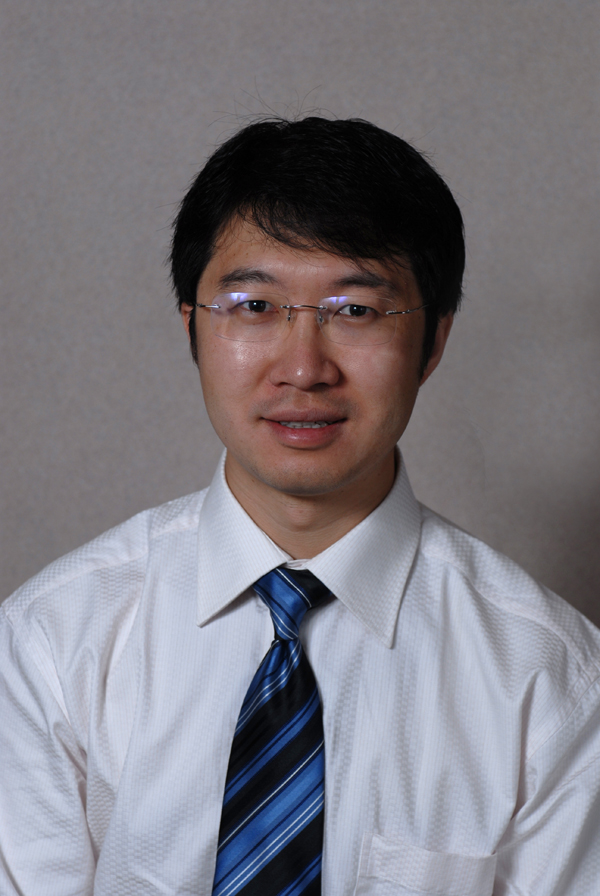}}]{Tony X. Han}
	received the B.S. degree with honors in Electrical Engineering Department and Special Gifted Class from Jiaotong University, Beijing, China in 1998, M.S. degree in electrical and computer engineering from the University of Rhode Island, RI, in 2002, and Ph.D degree in electrical and computer engineering from the University of Illinois at Urbana-Champaign, IL, in 2007. He then joined the Department of Electrical and Computer Engineering at the University of Missouri, Columbia, MO, in August 2007. Currently, he is the CEO of Jingchi.ai. His research interests include machine learning, computer vision, and unmanned vehicle.   
\end{IEEEbiography}
\end{document}